# Efficient Exact Inference in Planar Ising Models

**Nicol N. Schraudolph**
**Dmitry Kamenetsky**
*Research School of Information Sciences and Engineering*
*Australian National University* –and–
*NICTA, Locked Bag 8001*
*Canberra ACT 2601, Australia*

ARXIV@SCHRAUDOLPH.ORG
DMITRY.KAMENETSKY@NICTA.COM.AU

## Abstract

We give polynomial-time algorithms for the exact computation of lowest-energy (ground) states, worst margin violators, log partition functions, and marginal edge probabilities in certain binary undirected graphical models. Our approach provides an interesting alternative to the well-known graph cut paradigm in that it does not impose any submodularity constraints; instead we require planarity to establish a correspondence with perfect matchings (dimer coverings) in an expanded dual graph. We implement a unified framework while delegating complex but well-understood subproblems (planar embedding, maximum-weight perfect matching) to established algorithms for which efficient implementations are freely available. Unlike graph cut methods, we can perform penalized maximum-likelihood as well as maximum-margin parameter estimation in the associated conditional random fields (CRFs), and employ marginal posterior probabilities as well as *maximum a posteriori* (MAP) states for prediction. Maximum-margin CRF parameter estimation on image denoising and segmentation problems shows our approach to be efficient and effective. A C++ implementation is available from http://nic.schraudolph.org/isinf/.

## 1. Introduction

Undirected graphical models are a popular tool in machine learning; they represent real-valued energy functions of the form

$$E'(\boldsymbol{y}) := \sum_{i \in \mathcal{V}} E'_i(y_i) + \sum_{(i,j) \in \mathcal{E}} E'_{ij}(y_i, y_j), \tag{1}$$

where the terms in the first sum range over the nodes $\mathcal{V} = \{1, 2, \ldots n\}$, and those in the second sum over the edges $\mathcal{E} \subseteq \mathcal{V} \times \mathcal{V}$ of an undirected graph $G(\mathcal{V}, \mathcal{E})$.

The *junction tree* decomposition provides an efficient framework for exact statistical inference in graphs that are (or can be turned into) trees of small cliques. The resulting algorithms, however, require time exponential in the clique size, *i.e.*, the *treewidth* of the original graph. The treewidth of many graphs of practical interest is prohibitively large— for instance, it grows as $O(n)$ for an $n \times n$ square lattice. A large number of approximate inference techniques have been developed so as to deal with such graphs, such as pseudo-likelihood (Besag, 1986), mean field approximation, loopy belief propagation (Weiss, 2001; Yedidia et al., 2003), tree reweighting (Wainwright et al., 2003, 2005), and tree sampling (Hamze and de Freitas, 2004).

## 1.1 The Ising Model

Efficient *exact* inference is possible in certain graphical models with binary node labels. Here we focus on *Ising* models, whose energy functions have the form $E : \{0, 1\}^n \to \mathbb{R}$ with

$$E(\boldsymbol{y}) \ := \ \sum_{(i,j) \in \mathcal{E}} [y_i \neq y_j] \, E_{ij}, \tag{2}$$

where $[\cdot]$ denotes the indicator function, *i.e.*, the cost $E_{ij}$ is incurred only in those states $\boldsymbol{y}$ where $y_i$ and $y_j$ disagree. Compared to the general model (1) for binary nodes, (2) imposes two additional restrictions: zero node energies, and edge energies in the form of disagreement costs. At first glance these constraints look severe; for instance, such systems must obey the symmetry $E(\boldsymbol{y}) = E(\neg \boldsymbol{y})$, where $\neg$ denotes Boolean negation (ones' complement). However, it is well known (*e.g.*, Globerson and Jaakkola, 2007) that adding a single node makes the Ising model (2) as expressive as the general model (1) for binary variables:

**Theorem 1** *Every energy function of the form* (1) *over $n$ binary variables is equivalent to an Ising energy function of the form* (2) *over $n + 1$ variables, with the additional variable held constant.*

**Proof** by construction: Two energy functions are equivalent if they differ only by a constant. Without loss of generality, denote the additional variable $y_0$ and hold it constant at $y_0 := 0$. Given an energy function of the form (1), construct an Ising model with disagreement costs as follows:

1. For each node energy function $E_i'(y_i)$, add a disagreement cost of $E_{0i} := E_i'(1) - E_i'(0)$, as shown in Figure 1a. Note that in both states of $y_i$, the energy of the resulting Ising model is shifted relative to $E_i'(y_i)$ by the same constant amount, namely $E_i'(0)$:

| $y_i$ | general | Ising energy |
|-------|---------|--------------|
| 0 | $E_i'(0)$ | $0 = E_i'(0) - E_i'(0)$ |
| 1 | $E_i'(1)$ | $E_{0i} = E_i'(1) - E_i'(0)$ |

2. For each edge energy function $E_{ij}'(y_i, y_j)$, add the three disagreement cost terms

$$\begin{aligned}
E_{ij} &:= \tfrac{1}{2}[(E_{ij}'(0,1) + E_{ij}'(1,0)) - (E_{ij}'(0,0) + E_{ij}'(1,1))], \\
E_{0i} &:= E_{ij}'(1,0) - E_{ij}'(0,0) - E_{ij}, \quad \text{and} \\
E_{0j} &:= E_{ij}'(0,1) - E_{ij}'(0,0) - E_{ij},
\end{aligned} \tag{3}$$

as shown in Figure 1b. Note that for all states of $y_i$ and $y_j$, the energy of the resulting Ising model is shifted relative to $E_i'(y_i)$ by the same constant amount, namely $E_{ij}'(0,0)$:

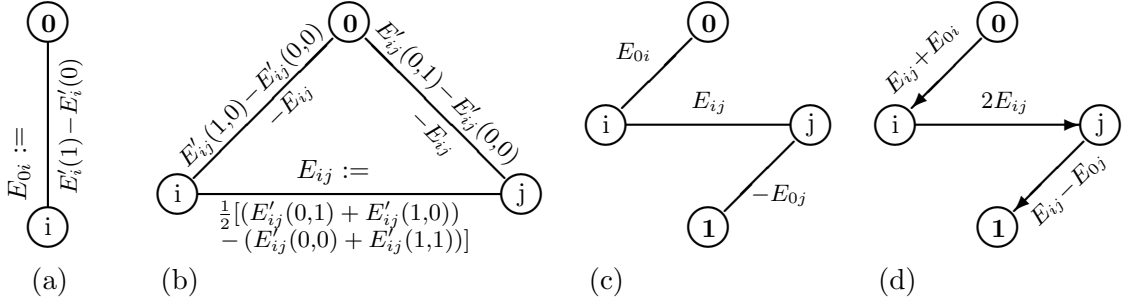

Figure 1: Equivalent Ising model (with disagreement costs) for a given (a) node energy $E_i'$, (b) edge energy $E_{ij}'$ in a binary graphical model; (c) equivalent submodular model if $E_{ij} > 0$ and $E_{0i} > 0$ but $E_{0j} < 0$; (d) equivalent directed model of Kolmogorov and Zabih (2004, Fig. 2d).

| $y_i$ | $y_j$ | general | Ising energy |
|---|---|---|---|
| 0 | 0 | $E_{ij}'(0,0)$ | $0 = E_{ij}'(0,0) - E_{ij}'(0,0)$ |
| 0 | 1 | $E_{ij}'(0,1)$ | $E_{0j} + E_{ij} = E_{ij}'(0,1) - E_{ij}'(0,0)$ |
| 1 | 0 | $E_{ij}'(1,0)$ | $E_{0i} + E_{ij} = E_{ij}'(1,0) - E_{ij}'(0,0)$ |
| 1 | 1 | $E_{ij}'(1,1)$ | $E_{0i} + E_{0j} = E_{ij}'(1,1) - E_{ij}'(0,0)$ |

Summing the above terms, the total *bias* of node $i$ (*i.e.*, its disagreement cost with the bias node) is

$$E_{0i} = E_i'(1) - E_i'(0) + \sum_{j:(i,j)\in\mathcal{E}} \left[E_{ij}'(1,0) - E_{ij}'(0,0) - E_{ij}\right]. \qquad (4)$$

This construction defines an Ising model whose energy in every configuration $\boldsymbol{y}$ is shifted, relative to that of the general model we started with, by the same constant amount, namely $E'(\boldsymbol{0})$:

$$\forall \boldsymbol{y} \in \{0,1\}^n : \quad E\left(\begin{bmatrix} 0 \\ \boldsymbol{y} \end{bmatrix}\right) = E'(\boldsymbol{y}) - \sum_{i\in\mathcal{V}} E_i'(0) - \sum_{(i,j)\in\mathcal{E}} E_{ij}'(0,0)$$
$$= E'(\boldsymbol{y}) - E'(\boldsymbol{0}). \qquad (5)$$

The two models' energy functions are therefore equivalent. ∎

Note how in the above construction the label symmetry $E(\boldsymbol{y}) = E(\neg\boldsymbol{y})$ of the plain Ising model (2) is conveniently broken by the introduction of a bias node, through the convention that $y_0 := 0$.

## 1.2 Energy Minimization via Graph Cuts

**Definition 2** *The* cut $\mathcal{C}$ *of a binary graphical model* $G(\mathcal{V}, \mathcal{E})$ *induced by state* $\boldsymbol{y} \in \{0,1\}^n$ *is the set* $\mathcal{C}(\boldsymbol{y}) := \{(i,j) \in \mathcal{E} : y_i \neq y_j\}$; *its* weight $|\mathcal{C}(\boldsymbol{y})|$ *is the sum of the weights of its edges.*

Any given state $\boldsymbol{y}$ partitions the nodes of a binary graphical model into two sets: those labeled '0', and those labeled '1'. The corresponding *graph cut* is the set of edges crossing the partition; since only they contribute disagreement costs to the Ising model (2), we have $\forall \boldsymbol{y} : |\mathcal{C}(\boldsymbol{y})| = E(\boldsymbol{y})$. The lowest-energy state of an Ising model therefore induces its minimum-weight cut. Conversely, the ground state of an Ising model (in absence of a bias node: up to label symmetry) can be determined from its minimum-weight cut via a simple (*e.g.*, depth-first) graph traversal (see Algorithm 1).

Minimum-weight cuts can be computed in polynomial time in graphs whose edge weights are all non-negative. Introducing one more node, with the constraint $y_{n+1} := 1$, allows us to construct an equivalent energy function by replacing each negatively weighted bias edge $E_{0i} < 0$ by an edge to the new node $n + 1$ with the positive weight $E_{i,n+1} := -E_{0i} > 0$ (Figure 1c). This still leaves us with the requirement that all non-bias edges be non-negative. This *submodularity* constraint implies that agreement between nodes must be locally preferable to disagreement — a severe limitation.

The now widespread use of graph cuts in machine learning to find lowest-energy configurations, in particular in image processing, was pioneered by Greig et al. (1989). Our construction (Figure 1c) differs from that of Kolmogorov and Zabih (2004) (Figure 1d) in that we do not employ the notion of *directed* edges. (In directed graphs, the weight of a cut is the sum of the weights of only those edges crossing the cut in a given direction.) We note that a more elaborate construction can give partial answers in graphs with some negative edge weights (Kolmogorov and Rother, 2007; Rother et al., 2007b), and that a sequence of *expansion moves* (energy minimizations in binary graphs) can efficiently yield an approximate answer for graphs with discrete but non-binary node labels (Boykov et al., 2001).

The remainder of this paper is organized as follows: Section 2 describes the planarity and connectivity conditions that our approach imposes upon the graphs, and how we handle them. Section 3 describes the construction of an expanded dual graph, and the algorithm for computing optimal (ground) states of the Ising model from it. The calculation of the partition function and marginal probabilities is dealt with in Section 4. These algorithms are then used in Section 5 to implement maximum-likelihood and maximum-margin parameter estimation in conditional random fields (CRFs). Section 6 describes our experiments using grid CRFs for image denoising and boundary detection, and Section 7 concludes with a discussion and outlook. We are making open source C++ code implementing our algorithms freely available for download from `http://nic.schraudolph.org/isinf/`.

## 2. Planarity and Connectivity

Unlike graph cut methods, the inference algorithms we will describe do not depend on submodularity; instead they require that the model graph be *planar*, and that a planar *embedding* be provided. They may also work only for *connected* graphs, and be the most efficient only for *biconnected* graphs. In this section we review these concepts, discuss their implications for our approach, and describe how to best handle them.

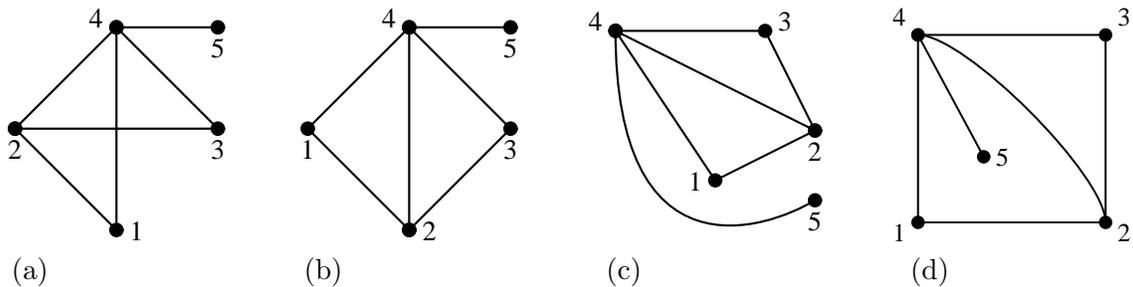

Figure 2: (a) a non-plane drawing of a planar graph; (b) a plane drawing of the same graph; (c) a different plane drawing of same graph, with the same planar embedding as (b); (d) a plane drawing of the same graph with a different planar embedding.

## 2.1 Embedding Planar Graphs

**Definition 3** *A graph is* planar *if it can be drawn in the plane $\mathbb{R}^2$ without edge intersections. The regions into which such a plane drawing partitions $\mathbb{R}^2$ are the* faces *of the drawing; the unbounded region is the* external face.

The operational nature of this definition would suggest that our algorithms must produce (or have access to) a plane drawing of the model graph. This is unsatisfactory in that such a drawing contains much information (such as the precise location of the vertices, and the exact shape of the edges) that we will not need. All we care about is the cyclic (say, clockwise) *ordering* of the edges incident upon each vertex. In topological graph theory, this is formalized in the notion of a *rotation system* (White and Beineke, 1978, p. 21f):

**Definition 4** *Let $G(\mathcal{V}, \mathcal{E})$ be an undirected, connected graph. For each vertex $i \in \mathcal{V}$, let $\mathcal{E}_i$ denote the set of edges in $\mathcal{E}$ incident upon $i$, considered as being oriented away from $i$, and let $\pi_i$ be a cyclic permutation of $\mathcal{E}_i$. A* rotation system *for $G$ is a set of permutations $\Pi = \{\pi_i : i \in \mathcal{V}\}$.*

To define the sets $\mathcal{E}_i$ of oriented edges more formally, construct the *directed* graph $G(\mathcal{V}, \mathcal{E}')$, where $\mathcal{E}'$ contains a pair of directed edges (known as *edgelets*) for each undirected edge in $\mathcal{E}$, that is, $(i, j) \in \mathcal{E}' \iff [(i, j) \in \mathcal{E} \lor (j, i) \in \mathcal{E}]$. Then $\mathcal{E}_i = \{(j, k) \in \mathcal{E}' : i = j\}$.

Rotation systems directly correspond to topological graph embeddings in orientable surfaces:

**Theorem 5** *Each rotation system determines an embedding of $G$ in some orientable surface $S$ such that $\forall i \in \mathcal{V}$, any edge $(i, j) \in \mathcal{E}_i$ is followed by $\pi_i(i, j)$ in (say) clockwise orientation, and such that the faces $\mathcal{F}$ of the embedding, given by the orbits of the mapping $(i, j) \rightarrow \pi_j(j, i)$, are 2-cells (topological disks).*

**Proof** see White and Beineke (1978, p. 22f). ∎

Note that while in graph visualisation "embedding" is often used as a synonym for "drawing", in modern topological graph theory it stands for "rotation system". We adopt the latter usage, which views embeddings as equivalence classes of graph drawings characterized by identical cyclic ordering of the edges incident upon each vertex. For instance, $\pi_4(4, 5) = (4, 3)$ in Figures 2b and 2c (same embedding) but $\pi_4(4, 5) = (4, 1)$ in Figure 2d (different embedding). A sample face in Figures 2b–2d is given by the orbit

$$(4, 1) \; \to \; \pi_1(1, 4) = (1, 2) \; \to \; \pi_2(2, 1) = (2, 4) \; \to \; \pi_4(4, 2) = (4, 1).$$

The *genus* $g$ of the embedding surface $S$ can be determined from the *Euler characteristic*

$$|\mathcal{V}| - |\mathcal{E}| + |\mathcal{F}| = 2 - 2g, \tag{6}$$

where $|\mathcal{F}|$ is found by counting the orbits of the rotation system, as described in Theorem 5. Since planar graphs are exactly those that can be embedded on a surface of genus $g = 0$ (a topological sphere), we arrive at a purely combinatorial definition of planarity:

**Definition 6** *A graph $G(\mathcal{V}, \mathcal{E})$ is* planar *iff it has a rotation system $\Pi$ producing exactly $2 + |\mathcal{E}| - |\mathcal{V}|$ orbits. Such a system is called a* planar embedding *of $G$, and $G(\mathcal{V}, \mathcal{E}, \Pi)$ is called a* plane graph.

Our inference algorithms require a plane graph as input. In certain domains (*e.g.*, when working with geographic information) a plane drawing of the graph (from which the corresponding embedding is readily determined) may be available. Where it is not, we employ the algorithm of Boyer and Myrvold (2004) which, given any connected graph $G$ as input, produces in linear time either a planar embedding for $G$ or a proof that $G$ is non-planar. Source code for this step is freely available (Boyer and Myrvold, 2004; Windsor, 2007).

## 2.2 The Planarity Constraint

In Section 1.1 we have mapped a general binary graphical model to an Ising model with an additional bias node; now we require that that Ising model be planar. What does that imply for the original, general model? If all nodes of the graph are to be connected to the bias node without violating planarity, the graph has to be *outerplanar*, *i.e.*, have a planar embedding in which all its nodes lie on the external face — a very severe restriction.

The situation improves, however, if we do not insist that all nodes be connected to the bias: If only a subset $\mathcal{B} \subset \mathcal{V}$ of nodes have non-zero bias (4), then the graph only needs to be $\mathcal{B}$-outerplanar, *i.e.*, have a planar embedding in which all nodes in $\mathcal{B}$ lie on the same face. Model selection may thus entail the step of picking the face of a suitably embedded planar Ising model whose nodes will be connected to the bias node. In image processing, for instance, where it is common to operate on a square grid of pixels, we can permit bias for all nodes on the perimeter of the grid, which borders the external face.

In general, a planar embedding which maximizes a weighted sum over the nodes bordering a given face can be found in linear time (Gutwenger and Mutzel, 2004); by setting node weights to some measure of their bias, such as the magnitude or square of $E_{0i}$ (4), we can thus efficiently obtain the planar Ising model closest (in that measure) to any given planar binary graphical model.

In contrast to submodularity, $\mathcal{B}$-outerplanarity is a structural constraint. This has the advantage that once a model obeying the constraint is selected, inference (*e.g.*, parameter estimation) can proceed via unconstrained methods (*e.g.*, optimization).

Finally, we note that all our algorithms can be extended to work for non-planar graphs as well. They then take time exponential in the genus of the embedding though still polynomial in the size of the graph; for graphs of low genus this may well be preferable to current approximative methods.

### 2.3 Connectivity

All algorithms in this paper assume that the graph $G(\mathcal{V}, \mathcal{E})$ is *connected*, *i.e.*, that $\mathcal{E}$ contains at least one path between any two nodes of $\mathcal{V}$. Where this is not the case, one can simply determine the connected components[1] of $G$ in linear time (Hopcroft and Tarjan, 1973), then invoke the algorithm in question separately on each of them. Since each component is unconditionally independent of all others (as they have no edges between them), the results can be trivially combined. Specifically,

- $G$ is planar iff all of its connected components are planar; any concatenation of a planar embedding for each connected component is a planar embedding of $G$.

- Any concatenation of a ground state for each connected component of $G$ is a ground state of $G$.

- The log partition function of $G$ is the sum of the log partition functions of its connected components.

- The edge marginal probabilities of $G$ are the concatenation of edge marginal probabilities of its connected components.

### 2.4 Biconnectivity

**Definition 7** *A graph is* biconnected *iff it is connected and does not have any articulation vertex. An* articulation vertex *is a vertex whose removal (along with any incident edges) disconnects the graph.*

Although the algorithms in this paper do not require $G(\mathcal{V}, \mathcal{E})$ to be biconnected, simpler and more efficient alternatives are applicable when this is not the case. As Figure 3 illustrates, any graph $G$ can be decomposed into a tree of edge-disjoint biconnected components[1] which overlap in the articulation vertices (of $G$) they share; this decomposition can be performed in linear time (Hopcroft and Tarjan, 1973). We make the following key observation:

**Theorem 8** *Let $\mathcal{E}_1, \mathcal{E}_2, \ldots, \mathcal{E}_n$ be the edge sets comprising the biconnected components of a graph $G(\mathcal{V}, \mathcal{E})$. The probability of cuts induced by the states of a Markov random field* (16)

---

1. A *component* of a graph with respect to a given property is a maximal subgraph that has the property.

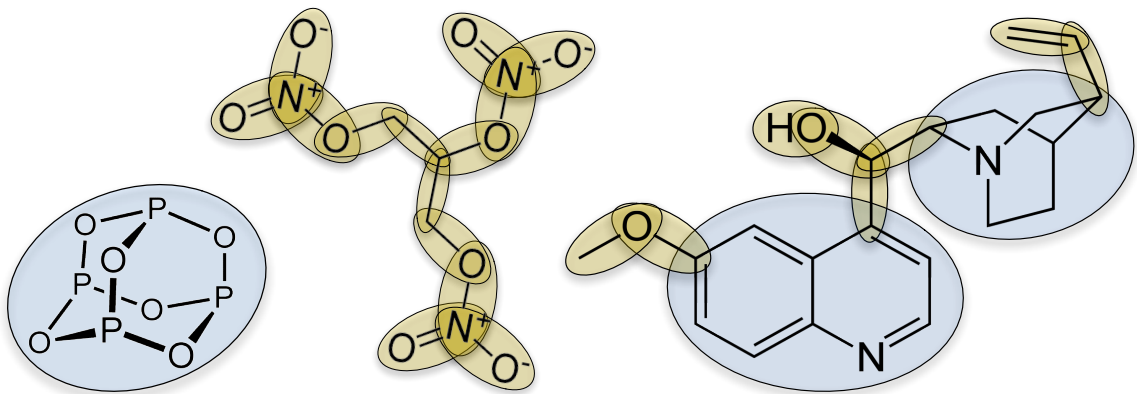

Figure 3: Skeletal chemical structures (images courtesy of Wikipedia) of phosphorus trioxide (bottom left), nitroglycerine (left of center), and quinine (right), with decomposition into a tree of biconnected components indicated (shaded ovals). Phosphorous trioxide is biconnected (*i.e.*, all one component); nitroglycerine is a tree (*i.e.*, has only trivial biconnected components).

*over an Ising model[2] (2) on $G$ factors into that of its biconnected components:*

$$\mathbb{P}[\mathcal{C}(\boldsymbol{y})] = \prod_{k=1}^{n} \mathbb{P}[\mathcal{C}(\boldsymbol{y}) \cap \mathcal{E}_k]. \tag{7}$$

**Proof** If $G(\mathcal{V}, \mathcal{E})$ is biconnected, $n = 1$ and $\mathcal{E}_1 = \mathcal{E}$, making Theorem 8 trivially true. Otherwise split $G$ into a biconnected component $G_1(\mathcal{V}_1, \mathcal{E}_1)$ which is a leaf of the decomposition tree, and the remainder $G'(\mathcal{V}', \mathcal{E}')$. It is always possible to find such a split. Because it is a leaf, $G_1$ connects to $G'$ through a single articulation vertex of $G$, which we call $v_1$. To summarize:

$$\mathcal{V}_1 \cup \mathcal{V}' = \mathcal{V}, \quad \mathcal{V}_1 \cap \mathcal{V}' = \{v_1\},$$
$$\mathcal{E}_1 \cup \mathcal{E}' = \mathcal{E}, \quad \mathcal{E}_1 \cap \mathcal{E}' = \emptyset, \quad G_1 \text{ biconnected.} \tag{8}$$

Let $\boldsymbol{y}_1$ and $\boldsymbol{y}'$ be the state $\boldsymbol{y}$ of the model restricted to vertices in $\mathcal{V}_1$ and $\mathcal{V}'$, respectively. By definition, $\boldsymbol{y}_1$ is independent of $\boldsymbol{y}'$ when both are conditioned on the state $y_{v_1}$ of the articulation vertex connecting them. We now make use of the label symmetry of Ising models, *resp.* the fact that it is broken by conditioning on a single node: $\mathbb{P}[\mathcal{C}(\boldsymbol{y})] = \mathbb{P}(\boldsymbol{y}|y_i)$ for any choice of $i$. We therefore have

$$\begin{aligned}
\mathbb{P}[\mathcal{C}(\boldsymbol{y})] &= \mathbb{P}(\boldsymbol{y}|y_{v_1}) = \mathbb{P}(\boldsymbol{y}_1, \boldsymbol{y}'|y_{v_1}) \\
&= \mathbb{P}(\boldsymbol{y}_1|y_{v_1}) \mathbb{P}(\boldsymbol{y}'|y_{v_1}) = \mathbb{P}[\mathcal{C}(\boldsymbol{y}) \cap \mathcal{E}_1] \mathbb{P}[\mathcal{C}(\boldsymbol{y}) \cap \mathcal{E}'].
\end{aligned} \tag{9}$$

---

2. Recall that we consider an Ising model to be an undirected graphical model with binary node states, no node potentials, edge potentials in the form of disagreement costs, and an optional bias node.

Recursively applying this argument to $G'$ yields Theorem 8.                                           ∎

The independence of biconnected components with respect to cuts $\mathcal{C}(\boldsymbol{y})$ stated by Theorem 8 may come as a surprise, given that the corresponding node states $\boldsymbol{y}$ do correlate through the articulation vertices. What happens is that label symmetry — specifically, the fact that $\mathcal{C}(\boldsymbol{y}) = \mathcal{C}(\neg\boldsymbol{y})$ — folds the state space so as to exactly cancel any moments between biconnected components. By decoupling their biconnected components, this facilitates efficient inference in graphs that are not biconnected themselves.

### 2.4.1 Ising Trees

Note in Figure 3 how edges which are not part of any cycle of $G$ form *trivial* biconnected components comprising only themselves and the two articulation vertices they connect. At the most extreme, a *tree* $T$ does not contain any cycles, hence consists entirely of trivial biconnected components (Figure 3, left of center). Theorem 8 then implies that each edge can be considered individually, making inference in an Ising tree $T(\mathcal{V}, \mathcal{E})$ very simple:

- $T$ is planar; any embedding of $T$ is a planar embedding.

- The minimum-weight cut of $T$ is the set $\mathcal{E}^- := \{(i, j) \in \mathcal{E} : E_{ij} < 0\}$.
  (Use Algorithm 1 to obtain the corresponding ground state.)

- The log partition function of $T$ is

$$\ln Z \;=\; \ln \prod_{(i,j) \in \mathcal{E}} (e^0 + e^{-E_{ij}}) \;=\; \sum_{(i,j) \in \mathcal{E}} \ln(1 + e^{-E_{ij}}). \tag{10}$$

- The marginal probability of any edge $(i, j)$ of $T$ is

$$\mathbb{P}[(i, j) \in \mathcal{C}] \;=\; \frac{e^{-E_{ij}}}{e^0 + e^{-E_{ij}}} \;=\; \frac{1}{1 + e^{E_{ij}}}. \tag{11}$$

### 2.4.2 The General Case

The most efficient way to employ the inference algorithms in this paper on graphs $G$ that are neither biconnected nor trees (*e.g.*, Figure 3, right) is to apply them to each nontrivial biconnected component of $G$ in turn, then use Theorem 8 to combine the results, along with the simple solutions given in Section 2.4.1 above for trivial biconnected components, into a result for the full graph. Letting $\mathcal{E}^T \subseteq \mathcal{E}$ denote the set of edges that belong to trivial biconnected components of $G$, we have:

- A planar embedding of $G$ is obtained by arbitrarily combining a planar embedding for each of its nontrivial biconnected components and the edges in $\mathcal{E}^T$.

- A minimum-weight cut of $G$ is the union between the edges in $\mathcal{E}^- \cap \mathcal{E}^T$ and a minimum-weight cut for each of $G$'s nontrivial biconnected components; Algorithm 1 can be used to obtain the corresponding ground state.

- The log partition function of $G$ is the sum of the log partition functions of its nontrivial biconnected components, plus $\sum_{(i,j)\in\mathcal{E}^{\mathcal{T}}} \ln(1 + e^{-E_{ij}})$.

- The marginal probability of an edge $(i,j) \in \mathcal{E}$ is $(1 + e^{E_{ij}})^{-1}$ if $(i,j) \in \mathcal{E}^T$, or computed (by the method of Section 4) from the biconnected component it belongs to.

This concludes our discussion of conditions on the Ising model graph $G$. In what follows, we assume that $G$ is connected and planar, and that a plane embedding is provided. We do not require that $G$ is biconnected, though where this is not the case, it is generally more efficient to decompose $G$ into biconnected components as discussed above.

## 3. Computing Ground States via Maximum-Weight Perfect Matching

**Definition 9** *A frustrated cycle $\mathcal{O} \subseteq \mathcal{E}$ of a graph $G(\mathcal{V}, \mathcal{E})$ with non-zero edge weights $E$ is a simple cycle whose product of edge weights is negative, i.e., $\prod_{(i,j)\in\mathcal{O}} E_{ij} < 0$. (A simple cycle is a closed path with no repeated edges or vertices.)*

A weighted graph is said to be frustrated if it contains any frustrated cycles. Note that trees can never be frustrated because they do not contain any cycles to begin with.

The lowest-energy (ground) state $\boldsymbol{y}^* := \operatorname{argmin}_{\boldsymbol{y}} E(\boldsymbol{y})$ of an *unfrustrated* Ising model is easily found via essentially the same method as in a tree (Section 2.4.1): paint nodes as you traverse the graph, flipping the binary color of your paintbrush whenever you traverse an edge with negative disagreement cost (as done by Algorithm 1 below when invoked on the cut $\mathcal{C} = \{(i,j) \in \mathcal{E} : E_{ij} < 0\}$). This cannot lead to a contradiction because by Definition 9 you will flip brush color an even number of times along any cycle in the graph, hence always end a cycle on the same color you started it with.

The presence of frustration unfortunately makes the computation of ground states much harder — in fact, it is known to be NP-hard in general (Barahona, 1982). As shown below, the ground state of a *planar* Ising model can be computed in polynomial time via maximum-weight perfect matching on an expanded dual of its embedded graph.

A relationship between the states of a planar Ising model and perfect matchings ("dimer coverings" to physicists) was first established by Kasteleyn (1961, 1963, 1967) and Fisher (1961, 1966). Perfect matchings in dual graph constructs were used by Bieche et al. (1980) and Barahona (1982) to compute Ising ground states; below we generalize a simpler construction for triangulated graphs due to Globerson and Jaakkola (2007). For rectangular lattices our approach reduces to the construction of Thomas and Middleton (2007), though their algorithm to compute ground states is somewhat less straightforward. Pardella and Liers (2008) apply this construction to very large square lattices (up to 3000×3000 nodes), and find it to be more efficient than earlier methods (Bieche et al., 1980; Barahona, 1982).

### 3.1 The Expanded Dual Graph

**Definition 10** *The dual $G^*(\mathcal{F}, \mathcal{E})$ of an embedded graph $G(\mathcal{V}, \mathcal{E}, \Pi)$ has a vertex for each face of $G$, with edges connecting vertices corresponding to faces that are adjacent (i.e., share an edge) in $G$.*

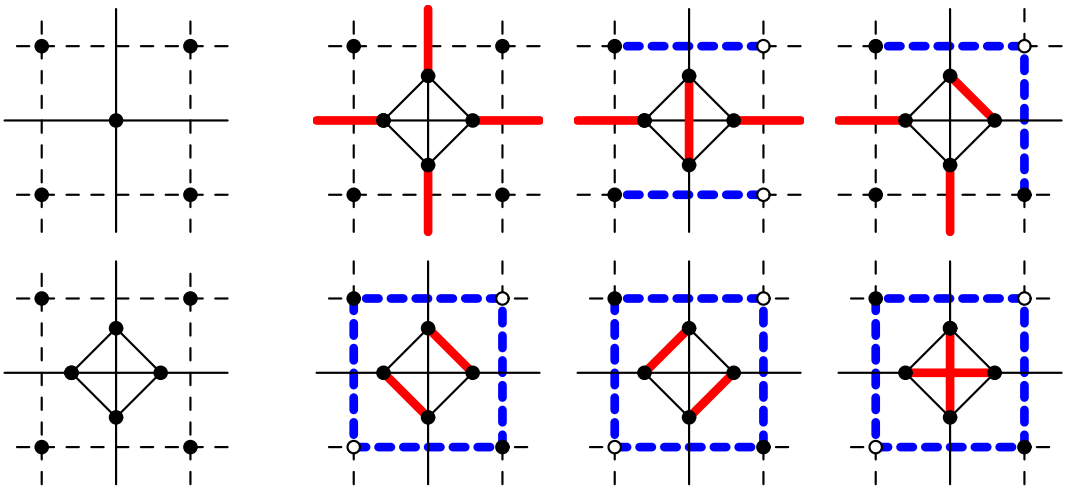

Figure 4: Left column: Square face of a plane graph (dashed lines) with its ordinary (top) *resp.* expanded (bottom) dual graph (solid lines). Other columns: Binary node states (open and filled circles) on the graph, induced cuts (bold blue dashes), and complementary perfect matchings (bold red lines) of the expanded dual.

Figure 4 (top left) shows the dual for a square face of our plane graph. Each edge of the dual crosses exactly one edge of the original graph; due to this one-to-one relationship we will consider the dual to have the *same* set of edges $\mathcal{E}$ (with the same energies) as the original. Since the nodes in the dual are different, however, we will (with some abuse of notation) use a single index for dual edges and their weights, corresponding to the index of the original edge in some arbitrary ordering of $\mathcal{E}$. Thus if $(i, j)$ is the $k^{\text{th}}$ edge in the (ordered) $\mathcal{E}$, its dual will have weight $E_k := E_{ij}$.

We now expand the dual graph by replacing each of its nodes with a $q$-clique, where $q$ is the degree of the node, as shown in Figure 4 (bottom left) for a dual node with degree $q = 4$, and in Figure 5 for an entire graph. The additional edges internal to each $q$-clique are given zero energy so as to leave the model unaffected. For large $q$ the introduction of these $O(q^2)$ internal edges slows down subsequent computations (solid line in Figure 8, left); this can be avoided by subdividing the offending $q$-gonal face of the model graph with chords (which are also given zero energy) before constructing the dual. Our implementation performs best when "octangulating" the graph, *i.e.*, splitting octagons off all faces with $q > 13$; this is more efficient than a full triangulation (Figure 8, left).

It is easily seen that the expanded dual has $2|\mathcal{E}|$ vertices, two for each edge in the original graph. We therefore give the two vertices connected by the dual of the $k^{\text{th}}$ edge in $\mathcal{E}$ the indices $2k - 1$ and $2k$ (*cf.* Section 4.3.1 and Figure 7). This consistent indexing scheme allows us to run the inference algorithms described in the remainder of this paper without explicitly constructing an expanded dual graph data structure.

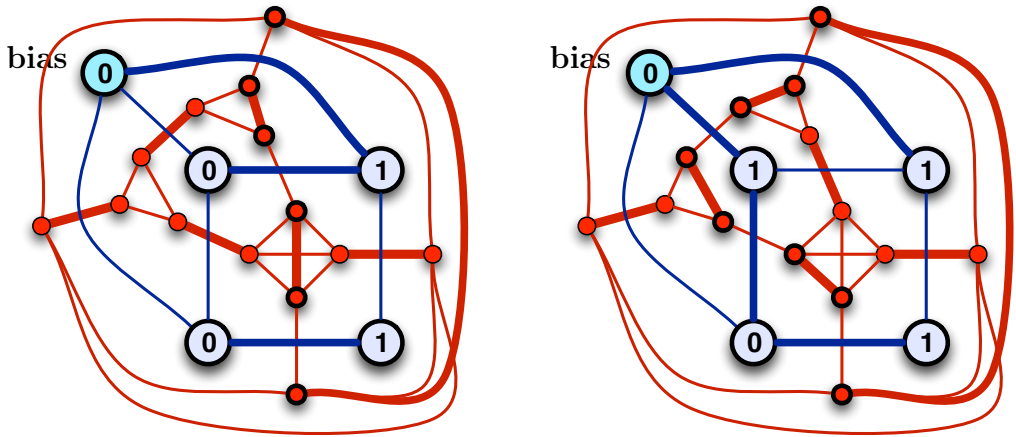

Figure 5: Example of a planar Ising model with five binary nodes (large, blue) including a constant-valued bias node (top left), and its expanded dual (small nodes, red), in two different states (left & right). The graph cut induced by the given state and its complementary perfect matching of the expanded dual are shown in bold, as are the nodes left unmatched by the complement $\mathcal{M}' := \mathcal{E}\backslash\mathcal{C}$ of the cut in the expanded dual.

### 3.2 Complementary Perfect Matchings

**Definition 11** *A perfect matching of a graph $G(\mathcal{V}, \mathcal{E})$ is a subset $\mathcal{M} \subseteq \mathcal{E}$ of edges wherein exactly one edge is incident upon each vertex: $\forall v \in \mathcal{V}, |v| = 1$ in $G(\mathcal{V}, \mathcal{M})$. Its weight $|\mathcal{M}|$ is the sum of the weights of its edges.*

Figure 5 shows two perfect matchings (in bold) of the nodes of the expanded dual of an Ising model. There is a complementary relationship between such perfect matchings and graph cuts in the original Ising model, characterized by the following two theorems. The reader may find it helpful to refer to Figure 5 while going through the proofs.

**Theorem 12** *For every cut $\mathcal{C}$ of an embedded graph $G(\mathcal{V}, \mathcal{E}, \Pi)$ there exists at least one (if $G$ is triangulated: exactly one) perfect matching $\mathcal{M}$ of its expanded dual complementary to $\mathcal{C}$, i.e., $\mathcal{E}\backslash\mathcal{M} = \mathcal{C}$.*

**Proof** By construction, the set $\mathcal{E}$ of edges of $G$ constitutes a perfect matching of the expanded dual. Any subset of $\mathcal{E}$ therefore is a (possibly partial) matching of the expanded dual. The complement $\mathcal{M}' := \mathcal{E}\backslash\mathcal{C}$ of a cut of $G$ is a subset of $\mathcal{E}$ and thus a matching of the expanded dual; it obeys $\mathcal{E}\backslash\mathcal{M}' = \mathcal{E}\backslash(\mathcal{E}\backslash\mathcal{C}) = \mathcal{C}$. The nodes that $\mathcal{M}'$ leaves *unmatched* in the expanded dual (circled bold in Figure 5) are those neighboring the edges $\mathcal{C}$ of the cut.

By definition, $\mathcal{C}$ intersects any cycle of $G$, and therefore also the perimeters of $G$'s faces $\mathcal{F}$, in an even number of edges. In each clique of the expanded dual, $\mathcal{M}'$ thus leaves an even number of nodes unmatched. It can therefore be completed into a perfect matching $\mathcal{M} \supseteq \mathcal{M}'$ using only edges interior to the cliques to pair up unmatched nodes. These edges have no counterpart in the original graph: $(\mathcal{M}\backslash\mathcal{M}') \cap \mathcal{E} = \emptyset$. We thus have

$$\mathcal{E}\backslash\mathcal{M} \;=\; \mathcal{E}\backslash[(\mathcal{M}\backslash\mathcal{M}') \cup \mathcal{M}'] \;=\; [\mathcal{E}\backslash(\mathcal{M}\backslash\mathcal{M}')]\backslash\mathcal{M}' \;=\; \mathcal{E}\backslash\mathcal{M}' \;=\; \mathcal{C}. \qquad (12)$$

In a 3-clique of the expanded dual, $\mathcal{M}'$ will leave two nodes unmatched or none at all; in either case there is only one way to complete the matching, by adding one edge *resp.* none at all. By construction, if $G$ is triangulated all cliques in its expanded dual are 3-cliques, and so $\mathcal{M}$ is unique. ∎

In other words, there exists a surjection from perfect matchings in the expanded dual of $G$ to cuts in $G$. Furthermore, since we have given edges interior to the cliques of the expanded dual zero energy (*i.e.*, $|\mathcal{M}\backslash\mathcal{M}'| = 0$), every perfect matching $\mathcal{M}$ complementary to a cut $\mathcal{C}$ of our Ising model (2) obeys the relation

$$|\mathcal{M}| + |\mathcal{C}| \;=\; |\mathcal{M}'| + |\mathcal{C}| \;=\; \sum_{(i,j)\in\mathcal{E}} E_{ij} \;=\; \text{const.} \qquad (13)$$

This means that instead of a minimum-weight cut in a graph we can look for a maximum-weight perfect matching in its expanded dual. But will that matching always be complementary to a cut? The following theorem shows that this is true for *plane* graphs:

**Theorem 13** *Every perfect matching $\mathcal{M}$ of the expanded dual of a plane graph $G(\mathcal{V}, \mathcal{E}, \Pi)$ is complementary to a cut $\mathcal{C}$ of $G$, i.e., $\mathcal{E}\backslash\mathcal{M} = \mathcal{C}$.*

**Proof** By definition, $\mathcal{E}\backslash\mathcal{M}$ is a cut of $G$ iff it intersects every cycle $\mathcal{O} \subseteq \mathcal{E}$ of $G$ an even number of times. This can be shown by induction over the faces of the embedding:

*Base case*—let $\mathcal{O} \subseteq \mathcal{E}$ be the perimeter of a face of the embedding, and consider the corresponding clique of the expanded dual: $\mathcal{M}$ matches an even number of nodes in the clique via interior edges; all other nodes must be matched by edges crossing $\mathcal{O}$. The complement of the matching in $G$ thus intersects $\mathcal{O}$ an even number of times:

$$|(\mathcal{E}\backslash\mathcal{M}) \cap \mathcal{O}| \equiv 0 \mod 2. \qquad (14)$$

*Induction*—let $\mathcal{O}_1, \mathcal{O}_2 \subseteq \mathcal{E}$ be cycles in $G$ obeying (14), and consider their *symmetric difference* $\mathcal{O}_1 \Delta \mathcal{O}_2 := (\mathcal{O}_1 \cup \mathcal{O}_2)\backslash(\mathcal{O}_1 \cap \mathcal{O}_2)$:

$$
\begin{aligned}
|(\mathcal{E}\backslash\mathcal{M}) \cap (\mathcal{O}_1\Delta\mathcal{O}_2)| \;&=\; |[(\mathcal{E}\backslash\mathcal{M}) \cap (\mathcal{O}_1 \cup \mathcal{O}_2)]\backslash(\mathcal{O}_1 \cap \mathcal{O}_2)| \\
&=\; |[(\mathcal{E}\backslash\mathcal{M}) \cap \mathcal{O}_1] \cup [(\mathcal{E}\backslash\mathcal{M}) \cap \mathcal{O}_2]| - |(\mathcal{E}\backslash\mathcal{M}) \cap (\mathcal{O}_1 \cap \mathcal{O}_2)| \\
&=\; |(\mathcal{E}\backslash\mathcal{M}) \cap \mathcal{O}_1| + |(\mathcal{E}\backslash\mathcal{M}) \cap \mathcal{O}_2| - 2|(\mathcal{E}\backslash\mathcal{M}) \cap (\mathcal{O}_1 \cap \mathcal{O}_2)| \\
&\equiv\; 0 + 0 - 2n \;\equiv\; 0 \mod 2. \quad (n \in \mathbb{N}) 
\end{aligned}
\qquad (15)
$$

We see that property (14) is preserved under composition of cycles via symmetric differences, and thus holds for all cycles that can be composed from face perimeters of the embedding of $G$ in this fashion.

All cycles in a *plane* graph $G$ are *contractible* on its embedding surface (a plane *resp.* sphere) because that surface has zero genus, and is therefore simply connected. All contractible cycles of $G$ can be constructed by composition of face perimeters via symmetric differences, thus have an even intersection with $\mathcal{E}\backslash\mathcal{M}$, which is therefore a cut. ∎

This is where planarity matters: Surfaces of non-zero genus are not simply connected, and thus non-plane graphs may contain non-contractible cycles (*e.g.*, encircling the hole of a torus). In the presence of frustration, our construction does not guarantee that the complement $\mathcal{E}\backslash\mathcal{M}$ of a perfect matching of the expanded dual contains an even number of edges along such cycles. For planar graphs, however, the above theorems allow us to leverage known polynomial-time algorithms for perfect matchings into inference methods for Ising models. This approach also works for non-planar Ising models that contain no frustrated non-contractible cycle.

We note that if all cliques of the expanded dual are of even size, there is also a direct (non-complementary) surjection from perfect matchings to cuts in the original graph. In contrast to our complementary map, the direct surjection requires the addition of dummy vertices into the expanded dual for faces of $G$ with odd perimeter so as to make the corresponding cliques even (Kasteleyn, 1963; Fisher, 1966; Liers and Pardella, 2008).

### 3.3 Computing the Ground State

The blossom-shrinking algorithm (Edmonds, 1965a,b) is a sophisticated method to efficiently compute the maximum-weight perfect matching of a graph. It can be implemented (Mehlhorn and Schäfer, 2002) to run in as little as $O(|\mathcal{E}|\,|\mathcal{V}|\log|\mathcal{V}|)$ time (Galil et al., 1986). Although the Blossom IV code we are using (Cook and Rohe, 1999) is asymptotically less efficient — $O(|\mathcal{E}|\,|\mathcal{V}|^2)$ — we have found it to be very fast in practice (Figure 8, left).

We can now efficiently compute the lowest-energy state of a planar Ising model as follows: Find a planar embedding of the model graph (Section 2.1), construct its expanded dual (Section 3.1), and run the blossom-shrinking algorithm on that to compute its maximum-weight perfect matching. Its complement in the original model is the minimum-weight graph cut (Section 3.2). We can identify the state which induces this cut via a depth-first graph traversal (Algorithm 1) that labels nodes as it encounters them, and checks for consistency on subsequent encounters. The traversal starts by labeling the bias node with its known state $y_0 := 0$.

## 4. Computing the Partition Function and Marginal Probabilities

A Markov random field (MRF) over our Ising model (2) models the distribution

$$\mathbb{P}(\boldsymbol{y}) \;=\; \tfrac{1}{Z}\,e^{-E(\boldsymbol{y})}, \quad \text{where} \quad Z := \sum_{\boldsymbol{y}} e^{-E(\boldsymbol{y})} \tag{16}$$

is the MRF's *partition function*. As it involves a summation over exponentially many states $\boldsymbol{y}$, calculating the partition function is generally intractable. For planar graphs, however, the generating function for perfect matchings can be calculated in polynomial time via the determinant of a skew-symmetric matrix (Kasteleyn, 1961, 1963, 1967; Fisher, 1961, 1966), which we call the *Kasteleyn matrix* $\boldsymbol{K}$. Due to the close relationship with graph cuts (Section 3.2) we can apply this method to calculate $Z$ in (16). We first convert a planar embedding of the Ising model graph into a Boolean "half-Kasteleyn" matrix $\boldsymbol{H}$, in four steps which will be elaborated below:

---

**Algorithm 1** Find State from Corresponding Graph Cut

---

| | |
|---:|:---|
| **Input:** | Ising model graph $G(\mathcal{V}, \mathcal{E})$, graph cut $\mathcal{C}(\boldsymbol{y}) \subseteq \mathcal{E}$ |
| 1. | $\forall i \in \{0, 1, 2, \ldots n\}: \; y_i :=$ unknown; |
| 2. | dfs_state(0, 0); |
| **Output:** | state vector $\boldsymbol{y}$ |

---

**procedure**    dfs_state($i \in \{0, 1, 2, \ldots n\}, s \in \{0, 1\}$)

      **if** $y_i =$ unknown **then**

          1. $y_i := s$;

          2. $\forall (i, j) \in \mathcal{E}_i :$

               **if** $(i, j) \in \mathcal{C}$ **then**

                    dfs_state($j, \neg s$);

               **else** dfs_state($j, s$);

      **else assert** $y_i = s$;

---

1. Section 4.1, Algorithm 2: *plane triangulate* the embedded graph so as to make the relationship between cuts and complementary perfect matchings a bijection (*cf.* Section 3.2);

2. Section 4.2, Algorithm 3: *orient* the edges of the graph such that the in-degree of every node is odd;

3. Section 4.3.3, Algorithm 4: *construct* the Boolean half-Kasteleyn matrix $\boldsymbol{H}$ from the oriented graph;

4. Section 4.4.3: *prefactor* the triangulation edges (added in Step 1) out of $\boldsymbol{H}$.

Our Step 2 simplifies equivalent operations in previous constructions (Kasteleyn, 1963, 1967; Fisher, 1966; Globerson and Jaakkola, 2007), Step 3 differs in that it only sets unit (*i.e.*, +1) entries in a Boolean matrix, and Step 4 can dramatically reduce the size of $\boldsymbol{H}$ for compact storage (as a bit matrix) and faster subsequent computations (Figure 8).

Edge disagreement costs do not enter into $\boldsymbol{H}$; they are only taken into account in a second phase, when the full Kasteleyn matrix $\boldsymbol{K}$ is constructed from $\boldsymbol{H}$ (Section 4.3.2). We can then factor $\boldsymbol{K}$ (Section 4.4) and compute the partition function from its determinant (Section 4.4.1; Kasteleyn, 1961; Fisher, 1961). This factorisation can also be used to invert $\boldsymbol{K}$, which is required to obtain *marginal probabilities* of disagreement on the edges of the model graph (Section 4.4.2).

In what follows, we elaborate in turn on the graph operations of plane triangulation (Section 4.1) and odd edge orientation (Section 4.2), and construction (Section 4.3) and factoring (Section 4.4) of the Kasteleyn matrix $\boldsymbol{K}$ *resp.* $\boldsymbol{H}$.

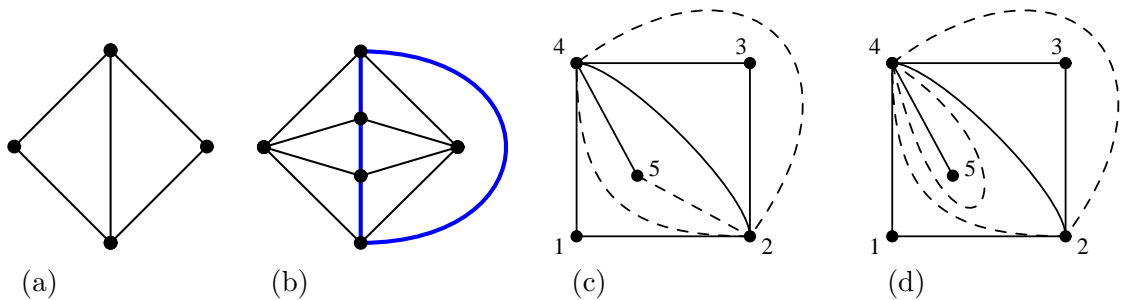

Figure 6: (a) A chordal graph whose external face is not triangulated; (b) a plane triangulated graph that has a 4-cycle (bold blue) without a chord; (c) proper, (d) improper plane triangulation (dashed) of the plane graph from Figure 2d.

## 4.1 Plane Triangulation

**Definition 14** *An embedded graph is* plane triangulated *iff it is biconnected and each of its faces (including the external face) is a triangle.*

Note that plane triangulation is *not* equivalent to making a graph *chordal*, though the latter process is sometimes also called "triangulation". For instance, the graph in Figure 6a is chordal but not plane triangulated because the external face is not triangular, while that in Figure 6b is plane triangulated but not chordal because it contains a 4-cycle (bold blue) that has no chord.

We can plane triangulate an embedded graph in linear time by traversing all of its faces, inserting chords as necessary as we go along (Algorithm 2). This may create multiple edges between the same two vertices, as shown in Figure 6c. Care must be taken when encountering singly connected components with a perimeter of length two, which could cause the insertion of a self-loop (see Figure 6d). Algorithm 2 detects and biconnects such components, as Definition 14 requires.

The insert_chord$(i, j, k)$ subroutine of Algorithm 2 not only updates $\mathcal{E}$, but also $\pi_i$ and $\pi_k$ so as to insert the new edge $(i, k)$ in its proper place in the rotation system. In order to leave the distribution (16) modeled by the graph unchanged, the new edge is given zero energy. Repeated traversals of the same face in Algorithm 2 can be avoided by the obvious use of "done" flags, omitted here for the sake of clarity.

Note that plane triangulation is not strictly necessary for the computation of partition function or marginal probabilities; it merely simplifies subsequent steps in the construction. Previously (Fisher, 1966; Globerson and Jaakkola, 2007) this convenience came at a computational price: the edges added during plane triangulation can make factoring and inversion of $\boldsymbol{K}$ (Section 4.4) significantly (up to 20 times) more expensive. We avoid this cost by removing the triangulation edges again before constructing $\boldsymbol{K}$, during prefactoring of the half-Kasteleyn bit matrix $\boldsymbol{H}$ (Section 4.4.3).

---

**Algorithm 2** Plane Triangulation

---

**Input:** plane graph $G(\mathcal{V}, \mathcal{E}, \Pi)$ with $|\mathcal{V}| \geq 3$

        $\forall i \in \mathcal{V}$ :

            $\forall (i,j) \in \mathcal{E}_i$ :

                1. $(j,k) := \pi_j(j,i);$

                2. $(k,l) := \pi_k(k,j);$

                3. **while** $l \neq i \ \vee \ \pi_l(l,k) \neq (l,j)$

                    (a) **if** $i = k$ **then**     (avoid self-loop)

                          $i := j;$

                          $j := k;$

                          $k := l;$

                          $(k,l) := \pi_k(k,j);$

                    (b) insert_chord$(i,j,k);$

                    (c) $i := k; \ j := l;$

                    (d) $(j,k) := \pi_j(j,i);$

                    (e) $(k,l) := \pi_k(k,j);$

**Output:** plane triangulated graph $G(\mathcal{V}, \mathcal{E}, \Pi)$

---

**procedure**   insert_chord$(i,j,k \in \mathcal{V})$

                1. $\mathcal{E} := \mathcal{E} \cup \{(i,k)\};$

                2. $\pi_k(k,i) := \pi_k(k,j);$

                3. $\pi_k(k,j) := (k,i);$

                4. $\pi_i(\pi_i^{-1}(i,j)) := (i,k);$

                5. $\pi_i(i,k) := (i,j);$

                6. $E_{ik} := 0;$

---

## 4.2 Odd Edge Orientation

To calculate the generating function for perfect matchings, the graph in question (namely, the expanded dual of our model graph) must be given a *clockwise odd orientation*.

**Definition 15** *An* orientation *of an undirected graph $G(\mathcal{V}, \mathcal{E})$ is a set $\mathcal{E}'$ of oriented edges with $|\mathcal{E}'| = |\mathcal{E}|$ such that $\forall (i,j) \in \mathcal{E}$, $\mathcal{E}'$ contains either $(i,j)$ or $(j,i)$.*

**Definition 16 (Kasteleyn, 1963)** *An orientation of an embedded graph is* clockwise odd *iff the number of edges oriented clockwise around each face (except possibly the external face) is odd.*

Consider Figure 7: by giving all interior edges of the 3-cliques of the expanded dual a clockwise orientation (small red arrows), we ensure that (a) the interior faces of the 3-cliques have clockwise odd orientation, and (b) all interior edges of the 3-cliques are oriented *counter*clockwise *wrt.* all faces exterior to the 3-cliques, hence do not affect the

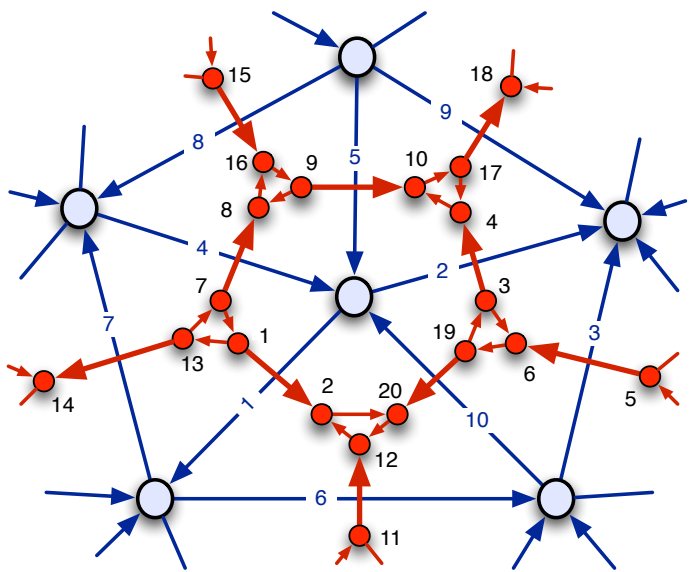

Figure 7: Clockwise odd orientation (Section 4.2) and indexing scheme (Section 4.3.1) for the expanded dual (red, small nodes) of the model graph (blue, large nodes).

latters' clockwise odd orientation status. It remains to consider the orientation of edges external to the 3-cliques (large red arrows). What does a clockwise odd orientation of these edges correspond to in the original model graph? To map these edges back into the model graph, *rotate* them clockwise by 90 degrees. A face with clockwise odd orientation of its perimeter in the dual thus maps to a vertex with an odd *in-degree*, *i.e.*, an odd number of edges oriented towards it. This facilitates a drastic simplification of this step in our construction:

*To establish a clockwise odd orientation of the expanded dual, simply orient the edges of the model graph such that all vertices, except possibly one, have an odd in-degree.*

Algorithm 3 achieves this in linear time by orienting edges appropriately upon return from a depth-first traversal of the graph. In contrast to earlier constructions (Kasteleyn, 1963; Fisher, 1966; Globerson and Jaakkola, 2007), it does not require following orbits around faces, and in fact does not refer to an embedding Π or dual graph $G^*$ at all.

Any vertex can be chosen to be the exceptional vertex $s$, since the choice of external face of a plane drawing is arbitrary — it is an artifact of the drawing, not an intrinsic property of the embedding: a planar graph embedded on a sphere has no external face.

## 4.3 Constructing the Kasteleyn Matrix

The *Kasteleyn matrix* $\boldsymbol{K}$ is a skew-symmetric, $2|\mathcal{E}| \times 2|\mathcal{E}|$ matrix constructed from the Ising model graph whose determinant is the square of the partition function. Our construction improves upon the work of Globerson and Jaakkola (2007) in a number of ways:

---

**Algorithm 3** Construct Odd Edge Orientation

---

| | |
|---|---|
| **Input:** | undirected graph $G(\mathcal{V}, \mathcal{E})$ |
| 1. | $\forall v \in \mathcal{V} : v.\text{visited} = \text{false};$ |
| 2. | pick arbitrary edge $(r, s) \in \mathcal{E};$ |
| 3. | $\mathcal{E}' := \{(r, s)\};$ |
| 4. | make_odd$(r, s);$ |
| **Output:** | orientation $\mathcal{E}'$ of $\mathcal{E} : \forall v \in \mathcal{V} \backslash \{s\},$ |
| | in-degree$(v) \equiv 1 \mod 2$ in $G(\mathcal{V}, \mathcal{E}')$ |

---

| | |
|---|---|
| **function** | make_odd: $(u, v \in \mathcal{V}) \rightarrow \{\text{true, false}\}$ |
| 1. | $\mathcal{E} := \mathcal{E} \backslash (u, v);$ |
| 2. | **if** $v.\text{visited}$ **then return** true; |
| 3. | $v.\text{visited} := \text{true};$ |
| 4. | odd := false; |
| 5. | $\forall w \in \mathcal{V} : \{v, w\} \in \mathcal{E}$ |
| |     **if** make_odd$(v, w)$ **then** |
| |     (a)  $\mathcal{E}' := \mathcal{E}' \cup \{(w, v)\};$ |
| |     (b)  odd := $\neg$ odd; |
| |     **else**  $\mathcal{E}' := \mathcal{E}' \cup \{(v, w)\};$ |
| 6. | **return** odd; |

---

- We employ an indexing scheme that obviates any need to refer to the expanded dual of the model graph (which we consequently never explicitly construct at all);

- We break construction of the Kasteleyn matrix into two phases, the first of which is invariant with respect to the model's disagreement costs;

- We make the "half-Kasteleyn" matrix $\boldsymbol{H}$ computed in the first phase very compact by *prefactoring* out the triangulation edges (see Section 4.4.3) and storing it as a bit matrix.

### 4.3.1 Indexing Scheme

Without loss of generality, let $\mathcal{E} = \{e_1, e_2, \ldots e_{|\mathcal{E}|}\}$. Note that the expanded dual has $2|\mathcal{E}|$ vertices, one lying to either side of every edge in the model graph. When viewing edge $e_k$ *along its direction in* $\mathcal{E}'$, we label the dual node to its right $2k - 1$ and that to its left $2k$; see Figure 7 for an example (*cf.* Section 3.1). One benefit of this scheme is that quantities relating to the edges of the model graph (as opposed to internal edges of the cliques of the expanded dual) will always be found on the superdiagonal of $\boldsymbol{K}$. We also notationally extend the rotation system from Section 2.1 to support this indexing scheme: $e_k = (i, j) \Longleftrightarrow e_{\pi_i(k)} = \pi_i(i, j).$

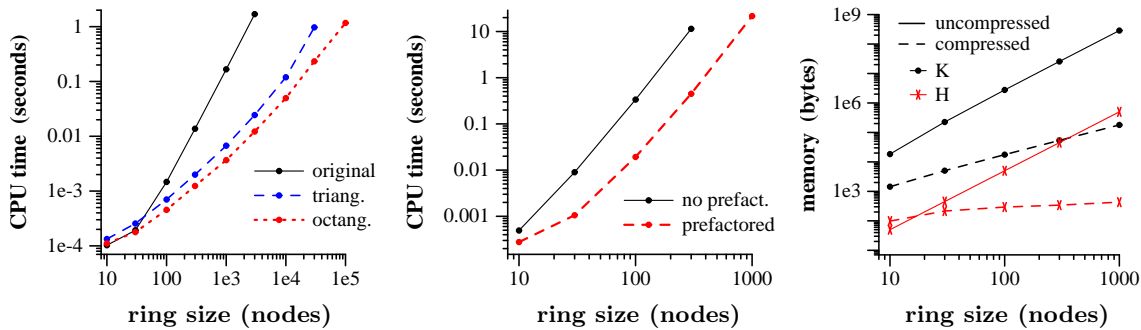

Figure 8: Cost of planar Ising inference methods on a ring graph, plotted against ring size. Left & center: CPU time on Apple MacBook with 2.2 GHz Intel Core2 Duo processor, averaged over 100 repetitions. Left: MAP state calculated via Blossom IV (Cook and Rohe, 1999) on original, triangulated, and octangulated ring. Center: marginal edge probabilities with *vs.* without prefactoring. Right: size of $\boldsymbol{K}$ (double precision, no prefactoring) *vs.* prefactored bit matrix $\boldsymbol{H}$, in uncompressed (solid lines) *vs.* compressed form (dashed lines), using row-compressed sparse matrix storage for $\boldsymbol{K}$ and `bzip2` compression for $\boldsymbol{H}$.

### 4.3.2 TWO-PHASE CONSTRUCTION

In a first phase we process the Ising model graph's *structure* into a Boolean "half-Kasteleyn" matrix $\boldsymbol{H}$ which does not yet include disagreement costs. For a given set of edge disagreement costs $E_k, k = \{1, 2, \dots |\mathcal{E}|\}$, we then build from $\boldsymbol{H}$ the conventional, real-valued Kasteleyn matrix $\boldsymbol{K}$ by adding the exponentiated disagreement costs along the superdiagonal and skew-symmetrizing:

1. $\boldsymbol{K} := \boldsymbol{H}$;

2. $\forall k \in \{1, 2, \dots, |\mathcal{E}|\} : \boldsymbol{K}_{2k-1, 2k} := \boldsymbol{K}_{2k-1, 2k} + e^{E_k}$;

3. $\boldsymbol{K} := \boldsymbol{K} - \boldsymbol{K}^\top$.

This two-phase approach holds a number of advantages:

- When working with a large number of isomorphic graphs (as we do in Section 6), the corresponding half-Kasteleyn matrix is identical for all of them, hence needs to be constructed just once.

- During maximum likelihood parameter estimation, partition function and/or marginals have to be recomputed many times for the same graph, with disagreement costs varying due to the ongoing adaptation of the model parameters. $\boldsymbol{H}$ remains valid when disagreement costs change, so we can compute it just once upfront, then re-use it in the parameter estimation loop.

- $\boldsymbol{H}$ can be stored very compactly as a prefactored bit matrix. As Figure 8 (right) shows, the uncompressed $\boldsymbol{H}$ can be several orders of magnitude smaller than the corresponding Kasteleyn matrix $\boldsymbol{K}$. Row-compressed sparse storage of $\boldsymbol{K}$ (which has exactly 3 non-zero entries in each row and column) is more efficient, but applying the `bzip2` compressor[3] to the prefactored bit matrix $\boldsymbol{H}$ yields by far the most compact storage format. Such memory efficiency becomes very important when working with large data sets of non-isomorphic graphs.

---

**Algorithm 4**    Construct Half-Kasteleyn Bit Matrix

---

**Input:**    oriented, embedded,

           triangulated graph $G(\mathcal{V}, \mathcal{E}', \Pi)$

   1.   $\boldsymbol{H} := \boldsymbol{0} \in \{0,1\}^{2|\mathcal{E}'| \times 2|\mathcal{E}'|}$

   2.   $\forall v \in \mathcal{V}$:

      (a)   $e_s :=$ any edge incident on $v$;

      (b)   **if** $e_s = (\cdot, v)$       (edge points to $v$)

            **then** $\alpha := 2s$;

            **else**   $\alpha := 2s - 1$;

      (c)   $i := \pi_v(s)$;

      (d)   **repeat**

            **if** $e_i = (\cdot, v)$      (edge points to $v$)

            **then**

                 $H_{2i-1,\alpha} := 1$;

                 $\alpha := 2i$;

                 **if** $e_i$ was created by Algorithm 2   (plane triangulation)

                     **then** $H_{2i-1,2i} := 1$;

            **else**

                 $H_{2i,\alpha} := 1$;

                 $\alpha := 2i - 1$;

              $i := \pi_v(i)$;

            **until** $i = \pi_v(s)$;

**Output:**   Half-Kasteleyn bit matrix $\boldsymbol{H}$

---

### 4.3.3 Algorithm for Constructing $\boldsymbol{H}$

Using the above indexing scheme, Algorithm 4 constructs $\boldsymbol{H}$ in linear time, cycling once through the edges incident upon each vertex of the model graph. It deviates from the classical construction of $\boldsymbol{K}$ (Kasteleyn, 1961; Fisher, 1961; Globerson and Jaakkola, 2007) in that it makes only positive entries, and only those corresponding to edges with zero disagreement cost, *i.e.*, added during plane triangulation or internal to the cliques of the

---

expanded dual. All such entries have the value $e^0 = 1$, making $\boldsymbol{H}$ a Boolean matrix, which can be stored compactly as a bit matrix.

### 4.4 Factoring Kasteleyn Matrices

Standard approaches such as LU-factorization can be used to factor the Kasteleyn matrix $\boldsymbol{K}$, but do not exploit its skew symmetry. Here we develop a numerically stable Cholesky-like factorization for even-sized skew-symmetric matrices, which can be used to factor $\boldsymbol{K}$ as well as to prefactor $\boldsymbol{H}$ (see below). Begin by writing the Kasteleyn matrix as

$$\boldsymbol{K} = \left[\begin{array}{cc|c} 0 & c & \boldsymbol{a}^\top \\ -c & 0 & \boldsymbol{b}^\top \\ \hline -\boldsymbol{a} & -\boldsymbol{b} & \boldsymbol{C} \end{array}\right], \tag{17}$$

for some scalar $c$, vectors $\boldsymbol{a}$ and $\boldsymbol{b}$, and a matrix $\boldsymbol{C}$ which is either empty or again of the same form. We factor (17) into (*cf.* Bunch, 1982; Benner et al., 2000)

$$\boldsymbol{K} = \left[\begin{array}{cc|c} 0 & -1 & \boldsymbol{0}^\top \\ 1 & 0 & \boldsymbol{0}^\top \\ \hline \boldsymbol{a}/c & \boldsymbol{b}/c & \boldsymbol{I} \end{array}\right] \left[\begin{array}{cc|c} 0 & c & \boldsymbol{0}^\top \\ -c & 0 & \boldsymbol{0}^\top \\ \hline \boldsymbol{0} & \boldsymbol{0} & \boldsymbol{C}' \end{array}\right] \left[\begin{array}{cc|c} 0 & 1 & \boldsymbol{a}^\top/c \\ -1 & 0 & \boldsymbol{b}^\top/c \\ \hline \boldsymbol{0} & \boldsymbol{0} & \boldsymbol{I} \end{array}\right], \tag{18}$$

where $\boldsymbol{C}'$ is the Schur complement

$$\boldsymbol{C}' := \boldsymbol{C} + (\boldsymbol{b}\boldsymbol{a}^\top - \boldsymbol{a}\boldsymbol{b}^\top)/c. \tag{19}$$

Iterated application of (18) to the Schur complement ultimately yields $\boldsymbol{K} = \boldsymbol{R}^\top \boldsymbol{J} \boldsymbol{R}$, where

$$\boldsymbol{R} := \left[\begin{array}{cc|cc|c} 0 & 1 & & & \boldsymbol{a}_1^\top/c_1 \\ -1 & 0 & & & \boldsymbol{b}_1^\top/c_1 \\ \hline 0 & 0 & 0 & 1 & \boldsymbol{a}_2^\top/c_2 \\ 0 & & -1 & 0 & \boldsymbol{b}_2^\top/c_2 \\ \hline \vdots & & \ddots & \ddots & \vdots \\ & & & 0 & 0 & 0 & 1 \\ 0 & & \cdots & & 0 & -1 & 0 \end{array}\right] \tag{20}$$

and

$$\boldsymbol{J} := \left[\begin{array}{cc|cc|c} 0 & c_1 & & 0 & \cdots & & 0 \\ -c_1 & 0 & & 0 & 0 & & \\ \hline 0 & 0 & 0 & c_2 & \ddots & & \vdots \\ 0 & & -c_2 & 0 & & & \\ & & & & 0 & & \\ \vdots & & \ddots & \ddots & & 0 & 0 \\ & & & & 0 & 0 & 0 & c_{|\mathcal{E}|} \\ 0 & & \cdots & & 0 & -c_{|\mathcal{E}|} & 0 \end{array}\right]. \tag{21}$$

To prevent small *pivots* $c_i$ from causing numerical instability, pivoting is required. Since Kasteleyn matrices have at least two entries of unit magnitude in each row and column, partial pivoting suffices.

### 4.4.1 Partition Function

The partition function for perfect matchings is $\sqrt{|\boldsymbol{K}|}$ (Kasteleyn, 1961; Fisher, 1961). Our factoring gives $|\boldsymbol{R}| = 1$ and $|\boldsymbol{J}| = \prod_i c_i^2$, so we have

$$\sqrt{|\boldsymbol{K}|} \;=\; \sqrt{|\boldsymbol{R}^\top| \, |\boldsymbol{J}| \, |\boldsymbol{R}|} \;=\; \sqrt{|\boldsymbol{J}|} \;=\; \prod_{i=1}^{|\mathcal{E}|} |c_i|. \tag{22}$$

Calculation of the product in (22) is prone to numerical overflow; this is easily avoided by working with logarithms instead. Using the complementary relationship (13) with graph cuts in planar Ising models, we obtain the log partition function for the latter as

$$\ln Z \;:=\; \ln \sum_{\boldsymbol{y}} e^{-\sum_{k \in \mathcal{C}(\boldsymbol{y})} E_k} \;=\; \ln \sum_{\boldsymbol{y}} e^{-(\sum_{k \in \mathcal{E}} E_k - \sum_{k \notin \mathcal{C}(\boldsymbol{y})} E_k)} \tag{23}$$

$$=\; \ln(e^{-\sum_{k \in \mathcal{E}} E_k} \sum_{\boldsymbol{y}} e^{\sum_{k \notin \mathcal{C}(\boldsymbol{y})} E_k}) \;=\; \ln \sqrt{|\boldsymbol{K}|} - \sum_{k \in \mathcal{E}} E_k \;=\; \sum_{i=1}^{|\mathcal{E}|} (\ln |c_i| - E_i).$$

### 4.4.2 Marginal Probabilities

The *marginal probability* of disagreement along an edge equals the negative gradient of the log partition function (23) with respect to the disagreement costs. Computing this involves the inverse of $\boldsymbol{K}$:

$$\mathbb{P}(k \in \mathcal{C}) \;=\; \sum_{\boldsymbol{y}:\, k \in \mathcal{C}(\boldsymbol{y})} \mathbb{P}(\boldsymbol{y}) \;=\; \tfrac{1}{Z} \sum_{\boldsymbol{y}:\, k \in \mathcal{C}(\boldsymbol{y})} e^{-E(\boldsymbol{y})} \;=\; -\frac{1}{Z} \frac{\partial Z}{\partial E_k} \;=\; -\frac{\partial \ln Z}{\partial E_k}$$

$$=\; 1 - \frac{1}{2|\boldsymbol{K}|} \frac{\partial |\boldsymbol{K}|}{\partial E_k} \;=\; 1 - \tfrac{1}{2} \operatorname{tr}\!\left(\boldsymbol{K}^{-1} \frac{\partial \boldsymbol{K}}{\partial E_k}\right) \;=\; 1 + \boldsymbol{K}^{-1}_{2k-1,2k}\, \boldsymbol{K}_{2k-1,2k}, \tag{24}$$

where tr denotes the matrix trace, and we have used the fact that $\boldsymbol{K}^{-1}$ is skew-symmetric as well. To invert $\boldsymbol{K}$, observe from (20) *resp.* (21) that $\boldsymbol{R}$ and $\boldsymbol{J}$ are essentially triangular *resp.* diagonal (simply swap rows $2k-1$ and $2k$, $k = 1, 2, \ldots |\mathcal{E}|$), and thus easily inverted. Then use $\boldsymbol{K}^{-1} = \boldsymbol{R}^{-1} \boldsymbol{J}^{-1} \boldsymbol{R}^{-\top}$ to obtain

$$[\boldsymbol{K}^{-1}]_{2k-1,2k} \;=\; \sum_{i=1}^{2|\mathcal{E}|} \sum_{j=1}^{2|\mathcal{E}|} [\boldsymbol{R}^{-1}]_{2k-1,i} [\boldsymbol{J}^{-1}]_{i,j} [\boldsymbol{R}^{-\top}]_{j,2k} \;=\; \frac{-1}{c_k} + \sum_{i=k+1}^{|\mathcal{E}|} d_{ik}, \tag{25}$$

$$\text{where} \quad d_{ik} \;:=\; \frac{[\boldsymbol{R}^{-1}]_{2k-1,2i}[\boldsymbol{R}^{-1}]_{2k,2i-1} - [\boldsymbol{R}^{-1}]_{2k-1,2i-1}[\boldsymbol{R}^{-1}]_{2k,2i}}{c_i}.$$

### 4.4.3 Prefactoring

Consider the rows and columns of $\boldsymbol{K}$ corresponding to an edge added during plane triangulation (Section 4.1). Re-order $\boldsymbol{K}$ to bring those rows and columns to the top left, so that they form the $\boldsymbol{a}, \boldsymbol{b}$, and $c$ of (17). Since the disagreement cost of a triangulation edge is zero, we now have a unity pivot: $c = e^0 = 1$. This has two advantageous consequences:

**Size reduction:** The unity pivot does not affect the value of the partition function. Since we are not interested in the marginal probability of triangulation edges (which after all are not part of the original model), we do not need $\boldsymbol{a}$ or $\boldsymbol{b}$ either, once we have computed the Schur complement (19). We can therefore discard the first two rows and first two columns of $\boldsymbol{K}$ after factoring (17). Factoring out all triangulation edges in this fashion reduces the size of $\boldsymbol{K}$ (*resp.* $\boldsymbol{R}$ and $\boldsymbol{J}$) to range only over the edges of the original Ising model graph. This reduces storage requirements and speeds up subsequent computation of the inverse (Figure 8, center).

**Boolean closure:** The unity pivot eliminates the division from the Schur complement (19); in fact we show below that applying (18) to prefactor $\boldsymbol{H} - \boldsymbol{H}^\top$ yields a Schur complement that can be expressed as $\boldsymbol{H}' - \boldsymbol{H}'^\top$, where $\boldsymbol{H}'$ is again a Boolean matrix. This closure property allows us to simply prefactor triangulation edges directly out of $\boldsymbol{H}$ without explicitly constructing $\boldsymbol{K}$.

Specifically, let $\boldsymbol{K} := \boldsymbol{H} - \boldsymbol{H}^\top$ for a half-Kasteleyn matrix $\boldsymbol{H}$ with elements in $\{0, 1\}$. Without loss of generality, assume that $\boldsymbol{H}$ and its transpose are *disjoint*, *i.e.*, have no non-zero element in common: $\boldsymbol{H} \odot \boldsymbol{H}^\top = \boldsymbol{0}$, where $\odot$ denotes Hadamard (element-wise) multiplication. Algorithm 4 respects this condition; violations would cancel in the construction of $\boldsymbol{K}$ anyway. Expressing $\boldsymbol{H}$ as

$$\boldsymbol{H} = \left[ \begin{array}{cc|c} 0 & 1 & \boldsymbol{a}_1^\top \\ 0 & 0 & \boldsymbol{b}_1^\top \\ \hline \boldsymbol{a}_2 & \boldsymbol{b}_2 & \boldsymbol{C}_1 \end{array} \right], \tag{26}$$

we can write $\boldsymbol{K} = \boldsymbol{H} - \boldsymbol{H}^\top$ as (17) with $\boldsymbol{a} = \boldsymbol{a}_1 - \boldsymbol{a}_2, \boldsymbol{b} = \boldsymbol{b}_1 - \boldsymbol{b}_2, c = 1$, and $\boldsymbol{C} = \boldsymbol{C}_1 - \boldsymbol{C}_1^\top$. The Schur complement (19) then becomes

$$\begin{aligned} \boldsymbol{C}' &= \boldsymbol{C}_1 - \boldsymbol{C}_1^\top + (\boldsymbol{b}_1 - \boldsymbol{b}_2)(\boldsymbol{a}_1 - \boldsymbol{a}_2)^\top - (\boldsymbol{a}_1 - \boldsymbol{a}_2)(\boldsymbol{b}_1 - \boldsymbol{b}_2)^\top \\ &= (\boldsymbol{C}_1 + \boldsymbol{b}_1\boldsymbol{a}_1^\top + \boldsymbol{b}_2\boldsymbol{a}_2^\top + \boldsymbol{a}_1\boldsymbol{b}_2^\top + \boldsymbol{a}_2\boldsymbol{b}_1^\top) - (\boldsymbol{C}_1^\top + \boldsymbol{a}_1\boldsymbol{b}_1^\top + \boldsymbol{a}_2\boldsymbol{b}_2^\top + \boldsymbol{b}_2\boldsymbol{a}_1^\top + \boldsymbol{b}_1\boldsymbol{a}_2^\top) \\ &=: \boldsymbol{H}' - \boldsymbol{H}'^\top, \end{aligned} \tag{27}$$

where

$$\boldsymbol{H}' = \boldsymbol{C}_1 + \boldsymbol{b}_1\boldsymbol{a}_1^\top + \boldsymbol{b}_2\boldsymbol{a}_2^\top + \boldsymbol{a}_1\boldsymbol{b}_2^\top + \boldsymbol{a}_2\boldsymbol{b}_1^\top. \tag{28}$$

It remains to show that all elements of $\boldsymbol{H}'$ are in $\{0, 1\}$. By definition of $\boldsymbol{H}$ (26), all elements of $\boldsymbol{C}_1, \boldsymbol{a}_1, \boldsymbol{a}_2, \boldsymbol{b}_1, \boldsymbol{b}_2$ are in $\{0, 1\}$, and by closure of multiplication in $\{0, 1\}$ so are their products. Thus an element of $\boldsymbol{H}'$ will be in $\{0, 1\}$ iff it is non-zero in at most one term on the right-hand side of (28), or (equivalently) iff all pairs formed from the five terms in question are disjoint. This can indeed be shown to be the case:

- Because $\boldsymbol{H} \odot \boldsymbol{H}^\top = \boldsymbol{0}$, we know that neither $\boldsymbol{a}_1$ and $\boldsymbol{a}_2$ nor $\boldsymbol{b}_1$ and $\boldsymbol{b}_2$ can have any non-zero elements in common, so $\boldsymbol{b}_1\boldsymbol{a}_1^\top$ and $\boldsymbol{b}_2\boldsymbol{a}_2^\top$ are disjoint, as are $\boldsymbol{a}_1\boldsymbol{b}_2^\top$ and $\boldsymbol{a}_2\boldsymbol{b}_1^\top$.

- By construction of $\boldsymbol{H}$ (Algorithm 4), $\boldsymbol{a}_1$ and $\boldsymbol{b}_1$ (*resp.* $\boldsymbol{a}_2$ and $\boldsymbol{b}_2$) can only have an element in common if the dual nodes on both sides of the corresponding edge are members of the same clique. This cannot happen because we explicitly ensure that the graph becomes biconnected during plane triangulation (Section 4.1), so that an edge cannot border the same face of the model graph on both sides. Thus all four outer products in (28) are pairwise disjoint.

- Finally, each outer product in (28) is disjoint from $\boldsymbol{C_1}$ as long as the edges being factored out do not form a cut of the model graph (*i.e.*, cycle of the dual). We are prefactoring only edges added during triangulation; for these to form a cut the model graph would have to have been disconnected prior to triangulation. This cannot happen here because we deal with (and eliminate) this possibility during earlier preprocessing (Section 2.3).

In summary, all five terms on the right-hand side of (28) are pairwise disjoint. Thus the Schur complement $\boldsymbol{H'}$ is a Boolean matrix as well, and can be computed from $\boldsymbol{H}$ (26) very efficiently by replacing the additions and multiplications in (28) with bitwise OR and AND operations, respectively. As long as further triangulation edges remain in $\boldsymbol{H'}$, we then set $\boldsymbol{H} := \boldsymbol{H'}$ and iteratively apply (26) and (28) so as to prefactor them out as well.

## 5. Application to CRF Parameter Estimation

Our algorithms can be applied to regularized maximum likelihood and maximum margin parameter estimation in conditional random fields (CRFs). In a standard planar Ising CRF, the disagreement costs in (2) are computed as $E_k := \boldsymbol{\theta}^\top \boldsymbol{x}_k$, *i.e.*, as inner products between local features (sufficient statistics) $\boldsymbol{x}_k$ of the modeled data at each edge $k$, and corresponding parameters $\boldsymbol{\theta}$ of the model. The *conditional* distribution $\mathbb{P}(\boldsymbol{y}|\boldsymbol{x}, \boldsymbol{\theta})$ (where $\boldsymbol{x}$ represents the union of all local features) is then modeled as a Markov random field (16).

### 5.1 Maximum Likelihood

Maximum-likelihood (ML) CRF parameter estimation seeks to minimize *wrt.* $\boldsymbol{\theta}$ the $L_2$-regularized negative log likelihood

$$
\begin{aligned}
L_{\mathrm{ML}}(\boldsymbol{\theta}) &:= \tfrac{1}{2}\lambda\|\boldsymbol{\theta}\|^2 - \ln \mathbb{P}(\boldsymbol{y}^*|\boldsymbol{x}, \boldsymbol{\theta}) \\
&= \tfrac{1}{2}\lambda\|\boldsymbol{\theta}\|^2 + E(\boldsymbol{y}^*|\boldsymbol{x}, \boldsymbol{\theta}) + \ln Z(\boldsymbol{\theta}|\boldsymbol{x})
\end{aligned}
\tag{29}
$$

of a given target labeling $\boldsymbol{y}^{*,4}$ with regularization parameter $\lambda$. This is a smooth, convex, non-negative objective that can be optimized via gradient methods such as LBFGS, either in conventional batch mode (Nocedal, 1980; Liu and Nocedal, 1989) or online (Schraudolph et al., 2007). The gradient of (29) with respect to the parameters $\boldsymbol{\theta}$ is given by

$$
\frac{\partial}{\partial \boldsymbol{\theta}} L_{\mathrm{ML}}(\boldsymbol{\theta}) = \lambda\boldsymbol{\theta} + \sum_{k\in\mathcal{E}} \Big([k \in \mathcal{C}(\boldsymbol{y}^*)] - \mathbb{P}(k \in \mathcal{C}(\boldsymbol{y}|\boldsymbol{x}, \boldsymbol{\theta}))\Big) \boldsymbol{x}_k.
\tag{30}
$$

The contribution of each edge $k$ to the gradient (30) is given by the product between its local features $\boldsymbol{x}_k$ and the difference between the indicator function for membership of $k$ in the cut induced by the target state $\boldsymbol{y}^*$ and the marginal probability of $k$ being contained in a cut, given $\boldsymbol{x}$ and $\boldsymbol{\theta}$. We compute the latter via the inverse of the Kasteleyn matrix (24).

---

4. For notational clarity we suppress here the fact that we are usually modeling a *collection* of data items; the objective function for such a set is simply the sum of objectives for each individual item in it.

## 5.2 Maximum Margin

For maximum-margin (MM) parameter estimation ([Taskar et al., 2004](#)) we instead minimize

$$L_{\mathrm{MM}}(\boldsymbol{\theta}) := \tfrac{1}{2}\lambda\|\boldsymbol{\theta}\|^2 + E(\boldsymbol{y}^*|\boldsymbol{x},\boldsymbol{\theta}) - \min_{\boldsymbol{y}} M(\boldsymbol{y}|\boldsymbol{y}^*,\boldsymbol{x},\boldsymbol{\theta}) \tag{31}$$

$$= \tfrac{1}{2}\lambda\|\boldsymbol{\theta}\|^2 + E(\boldsymbol{y}^*|\boldsymbol{x},\boldsymbol{\theta}) - E(\hat{\boldsymbol{y}}|\boldsymbol{x},\boldsymbol{\theta}) + d(\hat{\boldsymbol{y}}|\boldsymbol{y}^*),$$

where $\hat{\boldsymbol{y}} := \operatorname{argmin}_{\boldsymbol{y}} M(\boldsymbol{y}|\boldsymbol{y}^*,\boldsymbol{x},\boldsymbol{\theta})$ is the worst margin violator, *i.e.*, the state that minimizes, relative to a given target state $\boldsymbol{y}^*$,[4] the *margin energy*

$$M(\boldsymbol{y}|\boldsymbol{y}^*) := E(\boldsymbol{y}) - d(\boldsymbol{y}|\boldsymbol{y}^*), \tag{32}$$

where $d(\cdot|\cdot)$ is a measure of divergence in state space. If $d(\cdot|\cdot)$ is a weighted Hamming distance between induced cuts:

$$d(\boldsymbol{y}|\boldsymbol{y}^*) := \sum_{k\in\mathcal{E}} [[k\in\mathcal{C}(\boldsymbol{y})] \neq [k\in\mathcal{C}(\boldsymbol{y}^*)]]\, v_k, \tag{33}$$

where the $v_k > 0$ are constant weighting factors (in the simplest case: all ones) on the edges of our Ising model, then it is easily verified that the margin energy (32) is implemented (up to a shift that depends only on $\boldsymbol{y}^*$) by an isomorphic Ising model with disagreement costs

$$E_k + (2[k\in\mathcal{C}(\boldsymbol{y}^*)] - 1)\, v_k. \tag{34}$$

We can thus use our algorithm of Section 3.3 to efficiently find the worst margin violator.[5]

The maximum-margin objective (31) is convex but non-smooth; its gradient is

$$\frac{\partial}{\partial\boldsymbol{\theta}} L_{\mathrm{MM}}(\boldsymbol{\theta}) = \lambda\boldsymbol{\theta} + \sum_{k\in\mathcal{E}} \Big( [k\in\mathcal{C}(\boldsymbol{y}^*)] - [k\in\mathcal{C}(\hat{\boldsymbol{y}})] \Big)\, \boldsymbol{x}_k, \tag{35}$$

*i.e.*, local features $\boldsymbol{x}_k$ are multiplied by one of $\{-1, 0, 1\}$, depending on the membership of edge $k$ in the cuts induced by $\boldsymbol{y}^*$ and $\hat{\boldsymbol{y}}$, respectively. We can minimize (31) via bundle methods, such as the BT bundle trust algorithm ([Schramm and Zowe, 1992](#)), making use of the convenient lower bound $\forall\boldsymbol{\theta}: L_{\mathrm{MM}}(\boldsymbol{\theta}) \geq 0$, which holds because

$$\min_{\boldsymbol{y}} M(\boldsymbol{y}|\boldsymbol{y}^*,\boldsymbol{x},\boldsymbol{\theta}) \leq M(\boldsymbol{y}^*|\boldsymbol{y}^*,\boldsymbol{x},\boldsymbol{\theta}) = E(\boldsymbol{y}^*|\boldsymbol{x},\boldsymbol{\theta}) - d(\boldsymbol{y}^*|\boldsymbol{y}^*) = E(\boldsymbol{y}^*|\boldsymbol{x},\boldsymbol{\theta}). \tag{36}$$

## 6. Experiments

We now demonstrate the scalability of our approach to CRF parameter estimation (Section 5) on two simple computer vision problems: the synthetic binary image denoising task of [Kumar and Hebert](#) ([2004](#), [2006](#)), and the detection of segmentation boundaries in noisy masks from the GrabCut Ground Truth image segmentation database ([Rother et al., 2007a](#)).

---

5. [Thomas and Middleton](#) ([2007](#)) employ a similar approach to obtain the ground state from a given state $\boldsymbol{y}^*$ by setting up an isomorphic Ising model with disagreement costs $(1 - 2\,[k\in\mathcal{C}(\boldsymbol{y}^*)])E_k$.

### 6.1 Synthetic Binary Image Denoising

Kumar and Hebert (2004, 2006) developed an image denoising benchmark problem for binary random fields based on four hand-drawn $64 \times 64$ pixel images (Figure 9, top row). We created 50 instances of each image corrupted with pink noise, produced by convolving a white noise image (all pixels i.i.d. uniformly random) with a Gaussian density of one pixel standard deviation. Original and pink noise images were linearly mixed using signal-to-noise (S/N) amplitude ratios of $1 : n, n \in \mathbb{N}$. Figure 9 shows samples of the resulting noisy instances for S/N ratios of 1:5 (second row) and 1:6 (fourth row).

We then employed square grid planar Ising CRFs to denoise the images, with edge energies set to $E_{ij} := \langle [1, |x_i - x_j|], \boldsymbol{\theta} \rangle$, where $x_i$ is the pixel intensity at node $i$. The perimeter of the grid was connected to a bias node with constant input and label: $x_0 = y_0 = 0$. Bias edges had their own parameters, yielding CRFs with four parameters and up to (for a $64 \times 64$ grid) 4097 nodes and 8316 edges.

These systems were trained by maximum margin (MM) and maximum likelihood (ML) parameter estimation (Section 5) on the 50 noisy instances derived from the first image (Figure 9, left column) only. The gradient (35) *resp.* (30) was computed by adding the contributions from all 50 training instances to that of the regularizer, with $\lambda = 100$. We assessed the quality of the obtained parameters by determining (via the method of Section 3.3) the *maximum a posteriori* (MAP) states

$$\boldsymbol{y}^* := \underset{\boldsymbol{y}}{\operatorname{argmax}} \, \mathbb{P}(\boldsymbol{y}|\boldsymbol{x}, \boldsymbol{\theta}) = \underset{\boldsymbol{y}}{\operatorname{argmin}} \, E(\boldsymbol{y}|\boldsymbol{x}, \boldsymbol{\theta}) \tag{37}$$

of the trained CRF for all 150 noisy instances of the other three images. Interpreting these MAP states as attempted reconstructions of the original images, we then calculated the prediction error rates for both nodes and edges of the model.

All experiments were carried out on a Linux PC with 3 GB RAM and dual Intel Pentium 4 processors running at 3.6 GHz, each with 2 MB of level 2 cache.

#### 6.1.1 Noise Level

We first explored the limit of the ability of a full-size ($64 \times 64$) MM-trained Ising grid CRF to reconstruct the test images as the noise level increases. Rows 3 and 5 of Figure 9 show the reconstructions obtained from the noisy instances shown in rows 2 and 4, respectively. At low noise levels ($n < 5$) we obtained perfect reconstruction of the original images. At an S/N ratio of 1:5 the first subtle errors do creep in (Figure 9, third row), though less than 0.5% of the nodes (0.3% of the edges) are predicted incorrectly. At the 1:6 S/N ratio, these figures increase to 2.1% for nodes and 1.15% for edges, and the errors become far more noticeable (Figure 9, bottom row). For higher noise levels ($n > 6$) the reconstructions rapidly deteriorate as the noise finally overwhelms the signal. At these noise levels our human visual system was no longer able to accurately reconstruct the images either.

#### 6.1.2 Maximum Margin vs. Maximum Likelihood Parameter Estimation

Next we compared MM and ML parameter estimation at the S/N ratio of 1:6 (fourth row of Figure 9), where reconstruction begins to break down and any differences in performance should be evident. To make ML training computationally feasible, we subdivided each training image into 64 8×8 patches, then trained an 8×8 grid CRF on those patches. For MM training we used the full (64×64) images and model, as well as 32×32, 16×16, and 8×8 patches, so as to assess how this subdivision impacts the quality of the model. Testing always employed the MAP state of the full model on the full images.

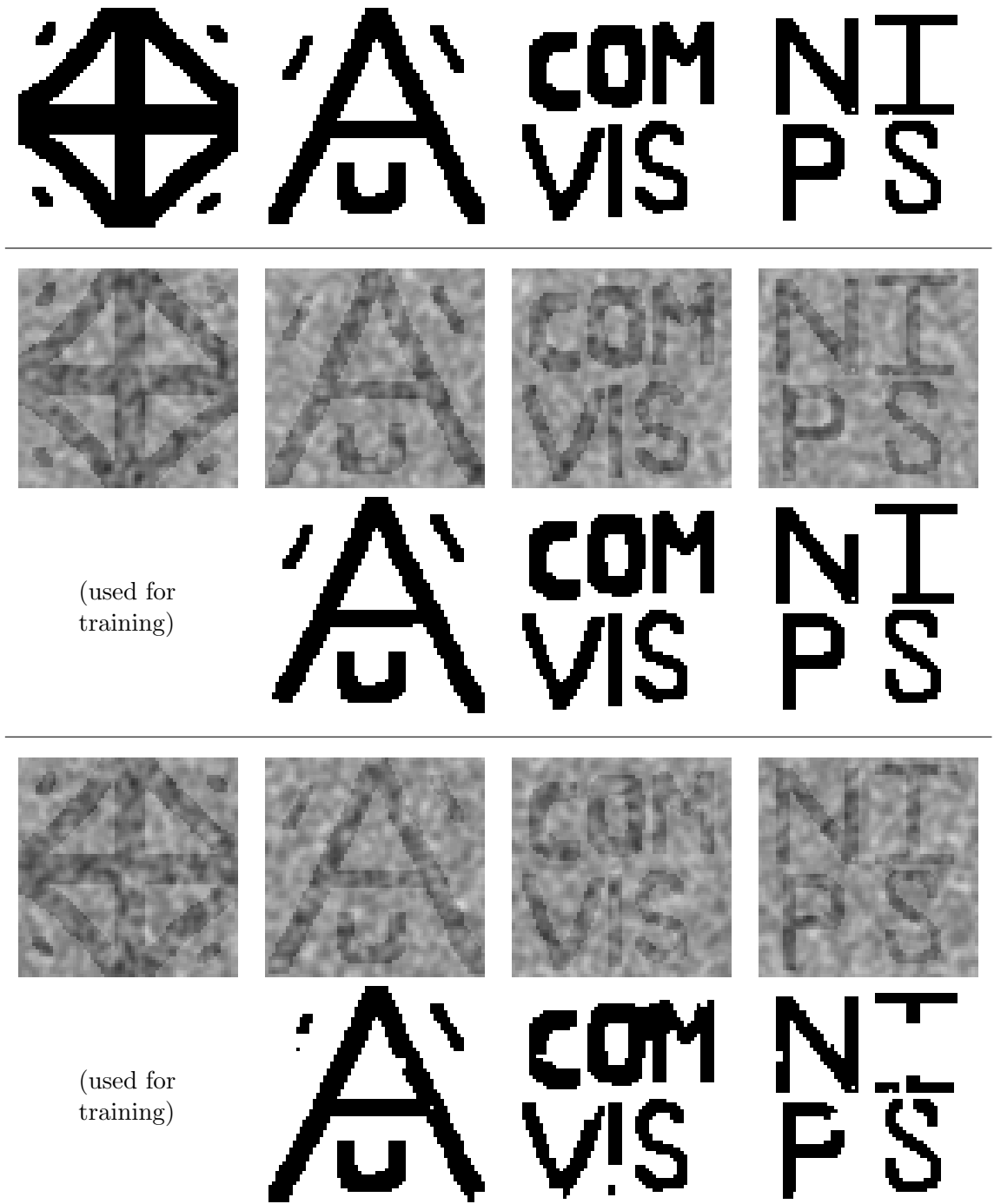

Figure 9: Denoising of binary images by maximum-margin training of planar Ising grids; from top to bottom: original images (Kumar and Hebert, 2004, 2006), images mixed with pink noise in a 1:5 ratio, reconstruction via MAP state of 64×64 Ising grid CRF from those noisy instances, images mixed with pink noise in a 1:6 ratio, and reconstruction from those noisy instances. Only the left-most image was used for training (max-margin CRF parameter estimation, λ = 100), the others for reconstruction.

Table 1: Performance comparison of parameter estimation methods on the denoising task; image reconstruction via MAP on the full (64×64) model and images. The minimum in each result column is boldfaced.

| Train Method | Patch Size | Train Time | Edge Error | Node Error |
|---|---|---|---|---|
| MM | $64 \times 64$ | $490.4\,\mathrm{s}$ | $1.15\,\%$ | $2.10\,\%$ |
| | $32 \times 32$ | $174.7\,\mathrm{s}$ | $1.16\,\%$ | $2.15\,\%$ |
| | $16 \times 16$ | $91.4\,\mathrm{s}$ | $1.12\,\%$ | $1.98\,\%$ |
| | $8 \times 8$ | $\mathbf{78.1\,s}$ | $\mathbf{1.10\,\%}$ | $\mathbf{1.83\,\%}$ |
| ML | | $5468.2\,\mathrm{s}$ | $1.11\,\%$ | $1.93\,\%$ |

Table 1 reports the edge and node errors obtained under each experimental condition. To assess statistical significance, we performed binomial pairwise comparison tests at a 95% confidence level against the null hypothesis that each of the two algorithms being compared has an equal (50%) chance of outperforming the other on a given test image.

We found no statistically significant difference here between 8×8 CRFs trained by MM *vs.* ML. However, ML training took 70 times as long to achieve this, so MM training is much preferable on computational grounds.

Counter to our expectations, the node and edge errors suggest that MM training actually works better on small (8×8 and 16×16) image patches. We surmise that this is because small patches have a relatively larger perimeter, leading to better training of the bias edges. Pairwise comparison tests, however, only found the node error for the 32×32 patch-trained CRF to be significantly worse than for the smaller patches; all other differences were below the significance threshold. What we can confidently state is that subdividing the images into small patches did not hurt performance, and yielded much shorter training times.

### 6.1.3 Maximum A Posteriori vs. Marginal Posterior Reconstruction

Fox and Nicholls (2001) have argued that the MAP state does not summarize well the information in the posterior distribution of an Ising model of noisy binary images, and proposed reconstruction via the *marginal posterior mode* (MPM) instead. For binary labels, the MPM is simply obtained by thresholding the marginal posterior node probabilities: $(\forall i \in \mathcal{V})\ y_i := [\mathbb{P}(y_i = 1 \,|\, \boldsymbol{x}, \boldsymbol{\theta}) > 0.5]$. In our Ising models, however, we have marginal posterior probabilities for edges (Section 4.4.2), and infer node states from graph cuts (Algorithm 1). Here implementing the MPM runs into a difficulty: since the edge set

$$\{k \in \mathcal{E} : \mathbb{P}(k \in \mathcal{C}(\boldsymbol{y}|\boldsymbol{x}, \boldsymbol{\theta})) > 0.5\} \tag{38}$$

may not be a cut of the model graph, hence may not unambiguously induce a node state. What we really need is the cut closest (in some given sense) to (38). For this purpose we

Table 2: Comparison of reconstruction methods on the denoising task, using the parameters of an MM-trained 8×8 CRF. The minimum in each result column is boldfaced; "test time" is the time required to reconstruct all 150 images.

| Test Method | Patch Size | Test Time | Edge Error | Node Error |
|---|---|---|---|---|
| MAP | 64×64 | 3.7 s | **1.10 %** | **1.83 %** |
| | 8×8 | **3.2 s** | 1.96 % | 5.21 % |
| M³P | | 397.5 s | 1.95 % | 3.32 % |

formulate the state $\boldsymbol{y}^+$ with the *maximal minimum marginal posterior* (M³P):

$$\boldsymbol{y}^+ := \operatorname*{argmax}_{\boldsymbol{y}'} \min_{k \in \mathcal{E}} \left\{ \begin{array}{ll} \mathbb{P}(k \in \mathcal{C}(\boldsymbol{y}|\boldsymbol{x}, \boldsymbol{\theta})) & \text{if } k \in \mathcal{C}(\boldsymbol{y}'), \\ 1 - \mathbb{P}(k \in \mathcal{C}(\boldsymbol{y}|\boldsymbol{x}, \boldsymbol{\theta})) & \text{otherwise.} \end{array} \right. \tag{39}$$

In other words, the M³P state (39) is induced by the cut whose edges (and those of its complement) have the largest minimum marginal probability. We can efficiently compute $\boldsymbol{y}^+$ as follows:

1. Find the minimum-weight spanning tree $\mathcal{E}^+ \subseteq \mathcal{E}$ of the model graph $G(\mathcal{V}, \mathcal{E})$ with edge weights $-|\mathbb{P}(k \in \mathcal{C}(\boldsymbol{y}|\boldsymbol{x}, \boldsymbol{\theta})) - 0.5|$. (This can be done in $O(|\mathcal{E}| \log |\mathcal{E}|)$ time.)

2. Run Algorithm 1 on $G(\mathcal{V}, \mathcal{E}^+)$ to find $\boldsymbol{y}^+$ as the state induced in the spanning tree by the MPM edge set (38).

Since $G(\mathcal{V}, \mathcal{E}^+)$ is a tree, it contains no cycles, and Algorithm 1 will therefore unambiguously identify the M³P state.

Table 2 lists the reconstruction errors achieved on the denoising task via MAP *vs.* M³P states, using the parameters of the MM-trained 8×8 CRF which gave the best performance in Section 6.1.2. (We obtained comparable results for ML training on 8×8 patches and MM training on full images.) Figure 10 shows representative sample reconstructions.

When reconstructing 8×8 patches, both MAP and M³P states appear to achieve virtually the same edge error. The binomial pairwise comparison test, however, reveals M³P to consistently outperform MAP here, even at a 99 % confidence level. This trend is confirmed by M³P's substantially lower node error. For comparison, an oracle that always picks the better of the two label-symmetric node states would yield a node error of 2.8 % here; the excess node error relative to that baseline is almost 5 times lower for M³P than for MAP. This does confirm the limitations of the MAP state for reconstruction, as argued by Fox and Nicholls (2001).

Since the M³P state (39) requires calculation of the edge marginals, however, it takes over two order of magnitude longer to obtain than the MAP state on an 8×8 patch, and is computationally infeasible for us to compute for the entire 64×64 image. The MAP state on the entire image clearly outperforms any reconstruction from 8×8 patches in terms of both edge and node error. The impressive scalability of the MAP state computation via blossom-shrinking (Section 3.3) thus in practice overrules here the theoretical advantage of reconstruction based on marginal probabilities.

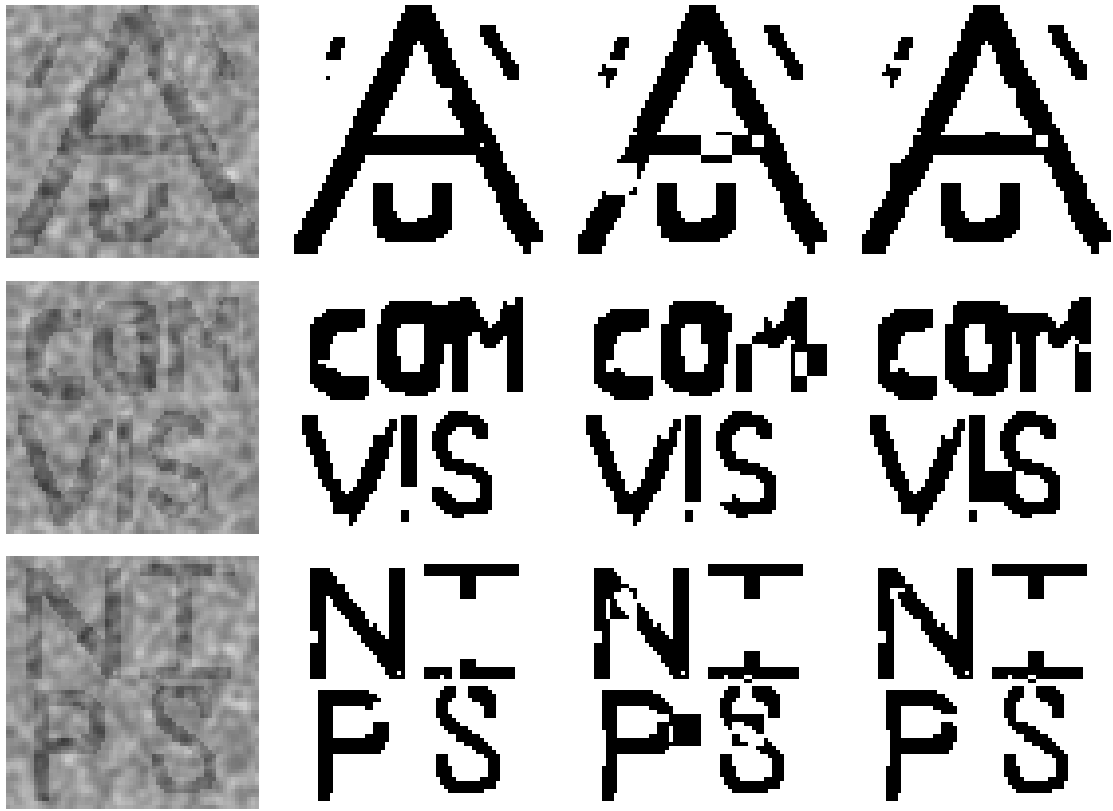

Figure 10: Image reconstructions on the denoising task; columns from left to right: Noisy 64×64 image, MAP reconstruction of full image, MAP reconstruction of 8×8 patches, and M³P reconstruction of 8×8 patches.

## 6.2 Boundary Detection

To further scale up our approach, we designed a simple boundary detection task based on images from the GrabCut Ground Truth image segmentation database (Rother et al., 2007a). We took 100×100 pixel subregions of images that depicted a segmentation boundary, and corrupted the segmentation mask with pink noise, produced by convolving a white noise image (all pixels i.i.d. uniformly random) with a Gaussian density with one pixel standard deviation.

We then employed a planar Ising model to recover the original boundary — namely, a 100×100 square grid with one additional edge pegged to a high energy, encoding prior knowledge that the bottom left and top right corners of the grid depict different regions. As before, the energy of the other edges was set to $E_{ij} := \langle [1, |x_i - x_j|], \boldsymbol{\theta} \rangle$, where $x_i$ is the pixel intensity at node $i$. We did not employ a bias node for this task, and simply set the regularization constant to $\lambda = 1$.

Note that this is a huge model: $10\,000$ nodes and $19\,801$ edges. Computing the partition function or marginals would require inverting a Kasteleyn matrix with over $1.5 \cdot 10^9$ entries; minimizing (29) is therefore computationally infeasible for us. Computing a ground state via the algorithm described in Section 3, by contrast, takes only 0.3 CPU seconds on an Apple MacBook with 2.2 GHz Intel Core2 Duo processor.[6] We can therefore efficiently minimize (31) to obtain the MM parameter vector $\boldsymbol{\theta}^*$, then compute the CRF's MAP (*i.e.*, ground) state for rapid prediction.

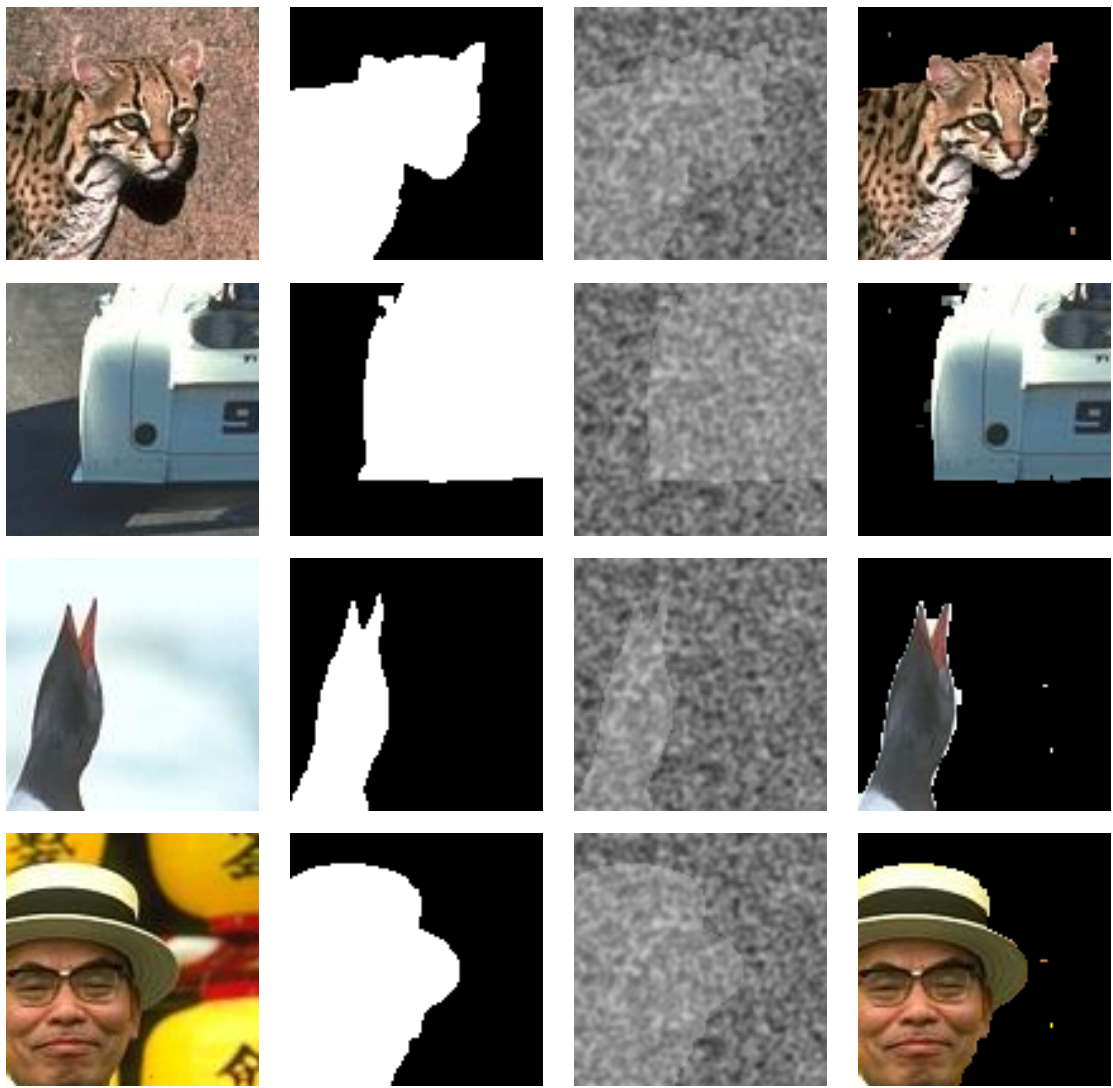

Figure 11: Boundary detection by maximum-margin training of planar Ising grids; columns from left to right: original image, original segmentation mask, noisy mask for S/N ratio of 1:8 (top two rows) *resp.* 1:7 (bottom two rows), and MAP segmentation obtained from the Ising grid CRF trained on the noisy mask.

As before, we titrated for the smallest S/N ratio of the form $1:n$ for which we obtained a good segmentation; depending on the image this occurred for $n = 7$ or 8. Figure 11 (right column) shows that at these noise levels our approach is capable of recovering the original segmentation boundary quite well, with less than 1% of nodes mislabeled. For S/N ratios of 1:9 and lower the system was unable to locate the boundary; for S/N ratios of 1:6 and higher we obtained perfect reconstruction. Again this corresponded closely to our human ability to visually locate the segmentation boundary accurately.

---

6. In scalability tests our code was able to compute the MAP state of a 400×400 grid in 0.85 CPU seconds.

## 7. Discussion and Outlook

We have proposed an alternative algorithmic framework for efficient exact inference in binary graphical models, which replaces the submodularity constraint of graph cut methods with a planarity constraint. Besides proving efficient and effective in first experiments, our approach opens up a number of interesting research directions to be explored:

Our algorithms can all be extended to nonplanar graphs, at a cost exponential in the genus of the embedding. We are currently developing these extensions, which may prove of great practical value for graphs that are "almost" planar; examples include road networks (where edge crossings arise from overpasses without on-ramps) and graphs describing the tertiary structure of proteins (Vishwanathan et al., 2007).

Our algorithms also provide a foundation for efforts to develop efficient *approximate* inference methods for nonplanar Ising models. Our method for calculating the ground state (Section 3) actually works for nonplanar graphs whose ground state does not contain frustrated non-contractible cycles. The QPBO graph cut method (Kolmogorov and Rother, 2007) finds ground states that do not contain any frustrated cycles, and otherwise yields a *partial* labeling. Can we likewise obtain a partial labeling of ground states with frustrated non-contractible cycles?

The existence of two distinct tractable frameworks for inference in binary graphical models implies a more powerful hybrid: Consider a graph each of whose biconnected components is either planar or submodular. In its entirety, this graph may be neither planar nor submodular, yet efficient exact inference in it is clearly possible by applying the appropriate framework to each component, then combining the results (Section 2.4.2). Can this hybrid approach be extended to cover less obvious situations?

### Acknowledgments

Earlier drafts of this paper have appeared on the arXiv preprint server (Schraudolph and Kamenetsky, 2008) and (in condensed form) at the 2008 NIPS conference (Schraudolph and Kamenetsky, 2009). We thank the anonymous NIPS reviewers for their valuable feedback.

NICTA is a national research institute with a charter to build Australia's pre-eminent centre of excellence for information and communications technology (ICT). NICTA is building capabilities in ICT research, research training and commercialisation for the generation of national benefit. NICTA is funded by the Australian Government as represented by the Department of Broadband, Communications and the Digital Economy and the Australian Research Council through the ICT Centre of Excellence program.

# References

Francisco Barahona. On the computational complexity of Ising spin glass models. *Journal of Physics A: Mathematical, Nuclear and General*, 15(10):3241–3253, 1982.

Peter Benner, Ralph Byers, Heike Fassbender, Volker Mehrmann, and David Watkins. Cholesky-like factorizations of skew-symmetric matrices. *Electronic Transactions on Numerical Analysis*, 11:85–93, 2000.

J. Besag. On the statistical analysis of dirty pictures. *Journal of the Royal Statistical Society B*, 48(3):259–302, 1986.

I. Bieche, R. Maynard, R. Rammal, and J. P. Uhry. On the ground states of the frustration model of a spin glass by a matching method of graph theory. *Journal of Physics A: Mathematical, Nuclear and General*, 13:2553–2576, 1980.

John M. Boyer and Wendy J. Myrvold. On the cutting edge: Simplified $O(n)$ planarity by edge addition. *Journal of Graph Algorithms and Applications*, 8(3):241–273, 2004. Reference implementation (C source code): http://jgaa.info/accepted/2004/BoyerMyrvold2004.8.3/planarity.zip.

Yuri Boykov, Olga Veksler, and Ramin Zabih. Fast approximate energy minimization via graph cuts. *IEEE Transactions on Pattern Analysis and Machine Intelligence*, 23(11): 1222–1239, 2001.

James R. Bunch. A note on the stable decomposition of skew-symmetric matrices. *Mathematics of Computation*, 38(158):475–479, April 1982.

William Cook and André Rohe. Computing minimum-weight perfect matchings. *INFORMS Journal on Computing*, 11(2):138–148, 1999. C source code available at http://www.isye.gatech.edu/~wcook/blossom4/.

Jack Edmonds. Maximum matching and a polyhedron with 0,1-vertices. *Journal of Research of the National Bureau of Standards*, 69B:125–130, 1965a.

Jack Edmonds. Paths, trees, and flowers. *Canadian Journal of Mathematics*, 17:449–467, 1965b.

Michael E. Fisher. Statistical mechanics of dimers on a plane lattice. *Physical Review*, 124 (6):1664–1672, 1961.

Michael E. Fisher. On the dimer solution of planar Ising models. *Journal of Mathematical Physics*, 7(10):1776–1781, 1966.

C. Fox and G. K. Nicholls. Exact map states and expectations from perfect sampling: Greig, Porteous and Seheult revisited. In A. Mohammad-Djafari, editor, *Bayesian Inference and Maximum Entropy Methods in Science and Engineering*, volume 568 of *American Institute of Physics Conference Series*, pages 252–263, 2001.

Zvi Galil, Silvio Micali, and Harold N. Gabow. An $O(EVlogV)$ algorithm for finding a maximal weighted matching in general graphs. *SIAM Journal of Computing*, 15:120–130, 1986.

Amir Globerson and Tommi Jaakkola. Approximate inference using planar graph decomposition. In B. Schölkopf, J. Platt, and T. Hofmann, editors, *Advances in Neural Information Processing Systems 19*, pages 473–480, Cambridge, MA, 2007. MIT Press.

D. M. Greig, B. T. Porteous, and A. H. Seheult. Exact maximum a posteriori estimation for binary images. *Journal of the Royal Statistical Society B*, 51(2):271–279, 1989.

Carsten Gutwenger and Petra Mutzel. Graph embedding with minimum depth and maximum external face. In G. Liotta, editor, *Graph Drawing 2003*, volume 2912 of *Lecture Notes in Computer Science*, pages 259–272. Springer Verlag, 2004.

F. Hamze and N. de Freitas. From fields to trees. In *Uncertainty in Artificial Intelligence (UAI)*, 2004.

John E. Hopcroft and Robert E. Tarjan. Algorithm 447: Efficient algorithms for graph manipulation. *Communications of the ACM*, 16(6):372–378, 1973.

Pieter W. Kasteleyn. The statistics of dimers on a lattice: I. the number of dimer arrangements on a quadratic lattice. *Physica*, 27(12):1209–1225, 1961.

Pieter W. Kasteleyn. Dimer statistics and phase transitions. *Journal of Mathematical Physics*, 4(2):287–293, 1963.

Pieter W. Kasteleyn. Graph theory and crystal physics. In Frank Harary, editor, *Graph Theory and Theoretical Physics*, chapter 2, pages 43–110. Academic Press, London and New York, 1967.

Vladimir Kolmogorov and Carsten Rother. Minimizing nonsubmodular functions with graph cuts – a review. *IEEE Trans. Pattern Analysis and Machine Intelligence*, 29(7):1274–1279, July 2007.

Vladimir Kolmogorov and Ramin Zabih. What energy functions can be minimized via graph cuts? *IEEE Trans. Pattern Analysis and Machine Intelligence*, 26(2):147–159, Feb. 2004.

Sanjiv Kumar and Martial Hebert. Discriminative fields for modeling spatial dependencies in natural images. In S. Thrun, L. Saul, and B. Schölkopf, editors, *Advances in Neural Information Processing Systems 16*, 2004.

Sanjiv Kumar and Martial Hebert. Discriminative random fields. *International Journal of Computer Vision*, 68(2):179–201, 2006. http://www-2.cs.cmu.edu/~skumar/DRF_IJCV.pdf.

Frauke Liers and Gregor Pardella. A simple max-cut algorithm for planar graphs. Technical Report zaik2008-579, Zentrum für Angewandte Informatik, Universität Köln, September 2008. http://www.zaik.uni-koeln.de/~paper/preprints.html?show=zaik2008-579.

Dong C. Liu and Jorge Nocedal. On the limited memory BFGS method for large scale optimization. *Mathematical Programming*, 45(3):503–528, 1989.

の

Kurt Mehlhorn and Guido Schäfer. Implementation of $O(nm \log n)$ weighted matchings in general graphs: The power of data structures. *ACM Journal of Experimental Algorithms*, 7:138–148, 2002. Article and C source code (requires LEDA 4.2 or higher) at http://www.jea.acm.org/2002/MehlhornMatching/.

Jorge Nocedal. Updating quasi-Newton matrices with limited storage. *Mathematics of Computation*, 35:773–782, 1980.

Gregor Pardella and Frauke Liers. Exact ground states of huge two-dimensional planar ising spin glasses. Technical Report 0801.3143, arXiv, January 2008. http://aps.arxiv.org/abs/0801.3143, to appear in *Phys Rev E*.

C. Rother, V. Kolmogorov, A. Blake, and M. Brown. GrabCut ground truth database. http://research.microsoft.com/vision/cambridge/i3l/segmentation/GrabCut.htm, 2007a.

Carsten Rother, Vladimir Kolmogorov, Victor Lempitsky, and Martin Szummer. Optimizing binary MRFs via extended roof duality. In *Proc. IEEE Conf. Computer Vision and Pattern Recognition*, Minneapolis, MN, June 2007b.

Helga Schramm and Jochem Zowe. A version of the bundle idea for minimizing a nonsmooth function: Conceptual idea, convergence analysis, numerical results. *SIAM J. Optimization*, 2:121–152, 1992.

Nicol N. Schraudolph and Dmitry Kamenetsky. Efficient exact inference in planar Ising models. Technical Report 0810.4401, arXiv, October 2008. http://aps.arxiv.org/abs/0810.4401.

Nicol N. Schraudolph and Dmitry Kamenetsky. Efficient exact inference in planar Ising models. In *Advances in Neural Information Processing Systems 21*, Cambridge, MA, to appear 2009. MIT Press.

Nicol N. Schraudolph, Jin Yu, and Simon Günter. A stochastic quasi-Newton method for online convex optimization. In *Proc. 11th Intl. Conf. Artificial Intelligence and Statistics (AIstats)*, San Juan, Puerto Rico, March 2007. Society for Artificial Intelligence and Statistics. http://nic.schraudolph.org/pubs/SchYuGue07.pdf.

B. Taskar, C. Guestrin, and D. Koller. Max-margin Markov networks. In S. Thrun, L. Saul, and B. Schölkopf, editors, *Advances in Neural Information Processing Systems 16*, pages 25–32, Cambridge, MA, 2004. MIT Press.

Creighton K. Thomas and A. Alan Middleton. Matching Kasteleyn cities for spin glass ground states. *Physical Review B*, 76(22):22406, 2007.

S. V. N. Vishwanathan, Karsten Borgwardt, and Nicol N. Schraudolph. Fast computation of graph kernels. In B. Schölkopf, J. Platt, and T. Hofmann, editors, *Advances in Neural Information Processing Systems 19*, Cambridge MA, 2007. MIT Press.

Martin J. Wainwright, Tommi S. Jaakkola, and Alan S. Willsky. Tree-based reparameterization framework for analysis of sum-product and related algorithms. *IEEE Transactions on Information Theory*, 49(5):1120–1146, 2003.

Martin J. Wainwright, Tommi S. Jaakkola, and Alan S. Willsky. A new class of upper bounds on the log partition function. *IEEE Transactions on Information Theory*, 51(7): 2313–2335, 2005.

Yair Weiss. Comparing the mean field method and belief propagation for approximate inference in MRFs. In David Saad and Manfred Opper, editors, *Advanced Mean Field Methods*. MIT Press, 2001.

Arthur T. White and Lowell W. Beineke. Topological graph theory. In Lowell W. Beineke and Robin J. Wilson, editors, *Selected Topics in Graph Theory*, chapter 2, pages 15–49. Academic Press, 1978.

Aaron Windsor. Planar graph functions for the boost graph library. C++ source code, boost file vault: `http://boost-consulting.com/vault/index.php?directory=Algorithms/graph`, 2007.

J.S. Yedidia, W.T. Freeman, and Y. Weiss. Understanding belief propagation and its generalizations. In *Exploring Artificial Intelligence in the New Millennium*, chapter 8, pages 239–236. Science & Technology Books, 2003.